\documentclass{article}

\usepackage{microtype}
\usepackage{graphicx}
\usepackage{subfigure}
\usepackage{booktabs} 

\usepackage{hyperref}

\newcommand{\ImageNetFinalInf}{85.5}
\newcommand{\ImageNetDiffInf}{22.8}
\newcommand{\ImageNetBestInf}{62.7}
\newcommand{\ImageNetStdInf}{8.87}

\newcommand{\ImageNetFinalTwo}{94.8}
\newcommand{\ImageNetDiffTwo}{31.8}
\newcommand{\ImageNetBestTwo}{63.0}
\newcommand{\ImageNetStdTwo}{1.16}

\newcommand{\cifarfinal}{51.4}
\newcommand{\cifarfinalstd}{0.41}
\newcommand{\cifarbest}{43.2}
\newcommand{\cifardiff}{8.2}
\newcommand{\cifarpreactvalidation}{46.9}
\newcommand{\cifarpreactbest}{46.7}
\newcommand{\cifarbesttrades}{42.3} 

\newcommand{\cifartwofinal}{31.1}
\newcommand{\cifartwobest}{28.4}
\newcommand{\cifartwodiff}{2.7}
\newcommand{\cifartwostd}{0.46}

\newcommand{\cifarwidefinal}{48.8}
\newcommand{\cifarwidebest}{41.8}

\newcommand{\tradesfinal}{50.6}

\newcommand{\tradesbest}{45.0}
\newcommand{\tradesdiff}{5.6}
\newcommand{\tradesreported}{43.4}
\newcommand{\tradesbesttrades}{44.1} 

\newcommand{\tradestwofinal}{58.2}
\newcommand{\tradestwostd}{0.66}
\newcommand{\tradestwobest}{53.6}
\newcommand{\tradestwodiff}{4.6}

\newcommand{\cifarlonebest}{48.6}
\newcommand{\cifarlonefinal}{53.0}
\newcommand{\cifarlonestd}{0.39}
\newcommand{\cifarlonediff}{4.4}

\newcommand{\cifarltwobest}{46.4}
\newcommand{\cifarltwofinal}{55.2}
\newcommand{\cifarltwostd}{0.4}

\newcommand{\cifarcutoutbest}{46.7}
\newcommand{\cifarcutoutfinal}{48.8}
\newcommand{\cifarcutoutdiff}{2.1}
\newcommand{\cifarcutoutstd}{0.79}

\newcommand{\cifarmixupbest}{46.3}
\newcommand{\cifarmixupfinal}{49.1}
\newcommand{\cifarmixupdiff}{2.8}
\newcommand{\cifarmixupstd}{1.32}

\newcommand{\cifarsemibest}{40.2}
\newcommand{\cifarsemifinal}{47.1}

\newcommand{\svhnfinal}{45.6}
\newcommand{\svhnbest}{39.0}
\newcommand{\svhndiff}{6.6}
\newcommand{\svhnstd}{0.40}

\newcommand{\svhntwofinal}{26.4}
\newcommand{\svhntwobest}{25.2}
\newcommand{\svhntwodiff}{1.2}
\newcommand{\svhntwostd}{0.27}

\newcommand{\cifarhunfinal}{78.6}
\newcommand{\cifarhunbest}{71.9}
\newcommand{\cifarhundiff}{6.7}
\newcommand{\cifarhunstd}{0.39}

\newcommand{\cifarhuntwofinal}{62.5}
\newcommand{\cifarhuntwobest}{56.8}
\newcommand{\cifarhuntwodiff}{5.7}
\newcommand{\cifarhuntwostd}{0.09}

\usepackage{todo}
\usepackage{amsmath}
\usepackage{multirow}
\usepackage{booktabs}
\usepackage{url}

\usepackage[accepted]{icml2020}

\icmltitlerunning{Overfitting in adversarially robust deep learning}

\begin{document}
\twocolumn[
\icmltitle{Overfitting in adversarially robust deep learning}



\icmlsetsymbol{equal}{*}

\begin{icmlauthorlist}
\icmlauthor{Leslie Rice}{equal,csd}
\icmlauthor{Eric Wong}{equal,mld}
\icmlauthor{J. Zico Kolter}{csd}
\end{icmlauthorlist}

\icmlaffiliation{csd}{Computer Science Department, Carnegie Mellon University, Pittsburgh PA, USA}
\icmlaffiliation{mld}{Machine Learning Department, Carnegie Mellon University, Pittsburgh PA, USA}

\icmlcorrespondingauthor{Leslie Rice}{larice@cs.cmu.edu}
\icmlcorrespondingauthor{Eric Wong}{ericwong@cs.cmu.edu}

\icmlkeywords{Machine Learning, ICML}

\vskip 0.3in
]



\printAffiliationsAndNotice{\icmlEqualContribution} 

\begin{abstract}
It is common practice in deep learning to use overparameterized networks and train for as long as possible; there are numerous studies that show, both theoretically and empirically, that such practices surprisingly do not unduly harm the generalization performance of the classifier. In this paper, we empirically study this phenomenon in the setting of \emph{adversarially trained deep networks}, which are trained to minimize the loss under worst-case adversarial perturbations. We find that overfitting to the training set does in fact harm robust performance to a very large degree in adversarially robust training across multiple datasets (SVHN, CIFAR-10, CIFAR-100, and ImageNet) and perturbation models ($\ell_\infty$ and $\ell_2$). Based upon this observed effect, we show that the performance gains of virtually all recent algorithmic improvements upon adversarial training can be matched by simply using early stopping.
We also show that effects such as the double descent curve \emph{do} still occur in adversarially trained models, yet fail to explain the observed overfitting.  Finally, we study several classical and modern deep learning remedies for overfitting, including regularization and data augmentation, and find that no approach in isolation improves significantly upon the gains achieved by early stopping. All code for reproducing the experiments as well as pretrained model weights and training logs can be found at \url{https://github.com/locuslab/robust_overfitting}.

\end{abstract}

\section{Introduction}

One of the surprising characteristics of deep learning is the relative \emph{lack} of overfitting seen in practice \citep{zhang2016understanding}. Deep learning models can often be trained to zero training error, effectively memorizing the training set, seemingly without causing any detrimental effects on the generalization performance. This phenomenon has been widely studied both from the theoretical \citep{neyshabur2017exploring} and empirical perspectives \citep{belkin2019reconciling}, and remains such a hallmark of deep learning practice that it is often taken for granted.

\begin{figure}[t]
\centering
\includegraphics{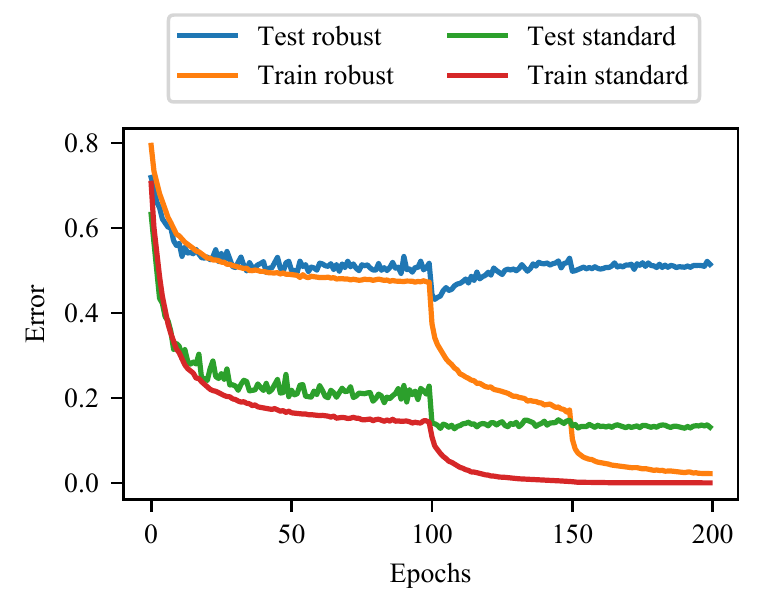}
\caption{The learning curves for a robustly trained model replicating the experiment done by \citet{madry2017towards} on CIFAR-10. The curves demonstrate ``robust overfitting''; shortly after the first learning rate decay the model momentarily attains \cifarbest{}\% robust error, and is actually more robust than the model at the end of training, which only attains \cifarfinal{}\% robust test error against a 10-step PGD adversary for $\ell_\infty$ radius of $\epsilon=8/255$. The learning rate is decayed at 100 and 150 epochs.}
\label{fig:cifar10_overfitting}
\end{figure}

In this paper, we consider the empirical question of overfitting in a similar, but slightly different domain: the setting of \emph{adversarial training} for robust networks. Adversarial training is a method for hardening classifiers against adversarial attacks, i.e. small perturbations to the input which can drastically change a classifier's predictions, that involves training the network on adversarially perturbed inputs instead of on clean data \citep{goodfellow2014explaining}.  It is generally regarded as one of the strongest empirical defenses against these attacks \citep{madry2017towards}.

A key finding of our paper is that, unlike in traditional deep learning, \emph{overfitting is a dominant phenomenon in adversarially robust training of deep networks.} That is, adversarially robust training has the property that, after a certain point, further training will continue to substantially decrease the robust training loss of the classifier, while increasing the robust test loss.  This is shown, for instance, in Figure \ref{fig:cifar10_overfitting} for adversarial training on CIFAR-10, where the robust test error dips immediately after the first learning rate decay, and only increases beyond this point.  We show that this phenomenon, which we refer to as ``robust overfitting'', can be observed on multiple datasets beyond CIFAR-10, such as SVHN, CIFAR-100, and ImageNet.

Motivated by this initial finding, we make several contributions in this paper to further study and diagnose this problem. First, we emphasize that virtually all the recent gains in adversarial performance from newer algorithms beyond simple projected gradient descent (PGD) based adversarial training \citep{mosbach2018logit, xie2019feature, yang2019me, zhang2019theoretically} can be attained by a much simpler approach: using early stopping. Specifically, by just using an earlier checkpoint, the robust performance of adversarially trained deep networks can be drastically improved, to the point where \emph{the original PGD-based adversarial training method can actually achieve the same robust performance as state-of-the-art methods}.
For example, vanilla PGD-based adversarial training \citep{madry2017towards} can achieve \cifarbest{}\% robust test error against a PGD adversary with $\ell_\infty$ radius 8/255 on CIFAR-10 when training is stopped early, on par with the \tradesreported{}\% robust test error reported by TRADES \citep{zhang2019theoretically} against the same adversary. This phenomenon is not unique to $\ell_\infty$ perturbations and is also seen in $\ell_2$ adversarial training. For instance, early stopping a CIFAR-10 model trained against an $\ell_2$ adversary with radius 128/255 can decrease the robust test error from \cifartwofinal\% to \cifartwobest\%.

Second, we study various empirical properties of overfitting for adversarially robust training and how they relate to standard training.
Since the effects of such overfitting appear closely tied to the learning rate schedule, we begin by investigating how changes to the learning rate schedule affect the prevalence of robust overfitting and its impacts on model performance.
We next explore how known connections between the hypothesis class size and generalization in deep networks translate to the robust setting, and show that the ``double descent'' generalization curves seen in standard training \citep{belkin2019reconciling} also hold for robust training \cite{nakkiran2019deep}. However, although this is used as a justification for the lack of overfitting in the standard setting, surprisingly, changing the hypothesis class size does not actually mitigate the robust overfitting that is observed during training.

Our final contribution is to investigate several techniques for preventing robust overfitting.
We first explore the effects of classic statistical approaches for combating overfitting beyond early stopping, namely explicit $\ell_1$ and $\ell_2$ regularization. We then study more modern approaches using data augmentation, including cutout \citep{devries2017improved}, mixup \citep{zhang2017mixup}, and semisupervised learning methods, which are known to empirically reduce overfitting in deep networks. Ultimately, while these methods can mitigate robust overfitting to varying degrees, when trained to convergence, \emph{we find that no other approach to combating robust overfitting performs better than simple early stopping}.
In fact, even combining regularization methods with early stopping tends to not significantly improve on early stopping alone. We find that the one exception is data augmentation with semi-supervised learning, where although the test performance can vary wildly even when training has converged, at select epochs it is possible to find a model with improved robust performance over simple early stopping. Code for reproducing all the experiments in this paper along with pretrained model weights and training logs can be found at \url{https://github.com/locuslab/robust_overfitting}.\footnote{Since there are over 75 models trained in this paper, we selected a subset of pretrained models to release (e.g. those which are for Wide ResNets since those take the most time to train, and can achieve the best performance in the paper)}

\section{Background and related work}
One of the first approaches to using adversarial training was with a single step gradient-based method for generating adversarial examples known as the fast gradient sign method (FGSM) \citep{goodfellow2014explaining}. The adversary was later extended to take multiple smaller steps, in a technique known as the basic iterative method \citep{kurakin2016adversarial}, and eventually reincorporated into adversarial training with random restarts, commonly referred to as projected gradient descent (PGD) adversarial training \citep{madry2017towards}. Further improvements to both the PGD adversary and the training procedure include incorporating momentum into the adversary \citep{dong2018boosting}, leveraging matrix estimation \citep{yang2019me}, logit pairing \citep{mosbach2018logit}, and feature denoising \citep{xie2019feature}. Most notably, \citet{zhang2019theoretically} proposed the method TRADES for adversarial training that balances the trade-off between standard and robust errors, and achieves state-of-the-art performance on several benchmarks.

Because PGD training is significantly more time consuming than standard training, several works have focused on improving the efficiency of adversarial training by reducing the computational complexity of calculating gradients and reducing the number of attack iterations \citep{shafahi2019adversarial, zhang2019you, wong2020fast}. Separate works have also expanded the general PGD adversarial training algorithm to different threat models including image transformations \citep{engstrom2017rotation, xiao2018spatially}, different distance metrics \citep{wong2019wasserstein}, and multiple threat models \citep{maini2019adversarial, tramer2019adversarial}.

Other adversarial defenses that have been proposed were not always successful, such as distillation \citep{papernot2016distillation, carlini2017towards} and detection of adversarial examples \citep{metzen2017detecting, feinman2017detecting, carlini2017adversarial, tao2018attacks, carlini2019ami}, which eventually were defeated by stronger attacks. Adversarial examples were also believed to be ineffective in the real world across different viewpoints \citep{lu2017no} until proven otherwise \citep{athalye2017synthesizing}, and a large number of adversarial defenses were shown to be relying on obfuscated gradients and ultimately rendered ineffective \citep{athalye2018obfuscated}, including thermometer encoding \citep{buckman2018thermometer} and various preprocessing techniques \citep{guo2017countering, song2017pixeldefend}.

Because many defenses were ``broken'' by stronger adversaries, a separate but related line of work has looked at generating certificates which can guarantee or prove robustness of the network output to norm-bounded adversarial perturbations. While not always scalable to large convolutional networks, methods for generating these robustness certificates range from using Satisfiability Modulo Theories (SMT) solvers \citep{ehlers2017formal, huang2017safety, katz2017reluplex} and mixed-integer linear programs \citep{tjeng2019evaluating} for exact certificates, to semi-definite programming (SDP) solvers for relaxed but still accurate certificates \citep{raghunathan2018certified, raghunathan2018semidefinite, fazlyab2019safety}.
Other methods focus on generating more tractable but relaxed certificates, which provide looser guarantees but can be optimized during training. These methods leverage techniques such as duality and linear programming \citep{wong2017provable, dvijotham2018dual, wong2018scaling, salman2019convex, zhang2019towards}, randomized smoothing \citep{cohen2019certified, lecuyer2019certified, salman2019provably}, distributional robustness \citep{sinha2017certifying}, abstract interpretations \citep{gehr2018ai2, mirman2018differentiable, singh2018fast}, and interval bound propagation \citep{gowal2018effectiveness}. Another approach is to use theoretically justified training heuristics \citep{croce2018provable, xiao2018training} which result in models which are verifiable by an independent certification method.

Highly relevant to this work are those that study the general problem of overfitting in machine learning. Both regularization \citep{friedman2001elements} and early stopping \citep{strand1974theory} have been well-studied in classical statistical settings to reduce overfitting and improve generalization, and connections between the two have been established in various settings such as in kernel boosting algorithms \citep{wei2017early}, least squares regression \citep{ali2018continuous}, and strongly convex problems \citep{suggala2018connecting}.
Although $\ell_2$ regularization (also known as weight decay) is commonly used for training deep networks \citep{krogh1992simple}, early stopping is less commonly used despite being studied as an implicit regularizer for controlling model complexity for neural networks at least 30 years ago \citep{morgan1990generalization}.\footnote{It is common practice in deep learning to save the best checkpoint which can be seen as early stopping. However, in the standard setting, the test loss tends to gradually improve over training, and so the best checkpoint tends to just select the best performance at the end of training, rather than stopping before training loss has converged.}
Indeed, it is now known that the standard bias-variance trade-off from classical statistical learning theory fails to explain why deep networks can generalize so well \citep{zhang2016understanding}. Consequently, it is now standard practice in many modern deep learning tasks to train for as long as possible and use large overparameterized models, since test set performance typically continues to improve past the point of dataset interpolation in what is known as ``double descent'' generalization \cite{belkin2019reconciling, nakkiran2019deep}. The generalization gap for robust deep networks has also been studied from a learning theoretic perspective in the context of data complexity \citep{schmidt2018adversarially} and Rademacher complexity \citep{yin2018rademacher}.

Also relevant to this work are methods specific to deep learning that empirically reduce overfitting and improve performance of deep networks. For example, Dropout is a commonly used stochastic regularization technique that randomly drops units and their connections from the network during training \citep{srivastava2014dropout} with the intent of preventing complex co-adaptations on the training data. Data augmentation is another technique frequently used when training deep networks that has been empirically shown to reduce overfitting. Cutout \citep{devries2017improved} is a form of data augmentation that randomly masks out a section of the input during training, which can be considered as augmenting the dataset with occlusions. Another technique known as mixup \citep{zhang2017mixup} trains on convex combinations of pairs of data points and their corresponding labels to encourage linear behavior in between data points.
Semi-supervised learning methods augment the dataset with unlabeled data, and have been shown to improve generalization when used in the adversarially robust setting \citep{carmon2019unlabeled, zhai2019adversarially, alayrac2019labels}.

\section{Adversarial training and robust overfitting}
In order to learn networks that are robust to adversarial examples, a commonly used method is adversarial training, which solves the following robust optimization problem
\begin{equation}
\min_\theta \sum_i \max_{\delta \in \Delta} \ell (f_\theta(x_i + \delta), y_i),
\end{equation}
where $f_\theta$ is a network with parameters $\theta$, $(x_i, y_i)$ is a training example, $\ell$ is the loss function, and $\Delta$ is the perturbation set. Typically the perturbation set $\Delta$ is chosen to be an $\ell_p$-norm ball (e.g.  $\ell_2$ and $\ell_\infty$ perturbations, which we consider in this paper), such that $\Delta=\{\delta: ||\delta||_p \leq \epsilon \}$ for $\epsilon > 0$. Adversarial training approximately solves the inner optimization problem, also known as the robust loss, using some adversarial attack method, typically with projected gradient descent (PGD), and then updates the model parameters $\theta$ using gradient descent \citep{madry2017towards}. For example, an $\ell_\infty$ PGD adversary would start at some random initial perturbation $\delta^{(0)}$
and iteratively adjust the perturbation with the following $\ell_\infty$ gradient steps while projecting back onto the $\ell_\infty$ ball with radius $\epsilon$:
\begin{equation}
\begin{split}
\tilde \delta &= \delta^{(t)} + \alpha \cdot \text{sign} \nabla_x \ell(f(x), y))\\
\delta^{(t+1)} &= \max(\min(\tilde \delta, \epsilon), -\epsilon)
\end{split}
\end{equation}
We denote error rates when attacked by a PGD adversary as the ``robust error'', and error rates on the clean, unperturbed data as ``standard error''.

\setlength{\tabcolsep}{4.0pt}
\begin{table}[t]
\caption{Robust performance showing the occurrence of robust overfitting across datasets and perturbation threat models. The ``best'' robust test error is the lowest test error observed during training. The final robust test error is averaged over the last five epochs. The difference between final and best robust test error indicates the degradation in robust performance during training.}
\begin{center}
\begin{small}
\begin{sc}
\label{tab:summary}
\begin{tabular}{lcrrrr}
\toprule
& & & \multicolumn{3}{c}{Robust test error ($\%$)} \\
Dataset & Norm & Radius &\multicolumn{1}{c}{Final} & Best & Diff\\
\midrule
\multirow{2}{*}{SVHN} & $\ell_\infty$ & $8/255$ & $\svhnfinal{}\pm \svhnstd{}$ & $\svhnbest{}$ & $\svhndiff{}$ \\
 & $\ell_2$ & $128/255$ & $\svhntwofinal{}\pm\svhntwostd{}$ & $\svhntwobest{}$ & $\svhntwodiff{}$ \\
 \midrule
 \multirow{2}{*}{CIFAR-10} & $\ell_\infty$ & $8/255$ & $\cifarfinal{}\pm \cifarfinalstd{}$ & $\cifarbest{}$ & $\cifardiff{}$\\
 & $\ell_2$ & $128/255$ & $\cifartwofinal{}\pm\cifartwostd{}$ & $\cifartwobest{}$ & $\cifartwodiff{}$ \\
 \midrule
 \multirow{2}{*}{CIFAR-100} & $\ell_\infty$ & $8/255$ & $\cifarhunfinal{}\pm\cifarhunstd{}$ & $\cifarhunbest{}$ & $\cifarhundiff{}$ \\
  & $\ell_2$ & $128/255$& $\cifarhuntwofinal{}\pm\cifarhuntwostd{}$ & $\cifarhuntwobest{}$ & $\cifarhuntwodiff{}$ \\
  \midrule
  \multirow{2}{*}{ImageNet} & $\ell_\infty$ & $4/255$ & $\ImageNetFinalInf{}\pm\ImageNetStdInf{}$ & $\ImageNetBestInf{}$ & $\ImageNetDiffInf{}$ \\
  & $\ell_2$ & $76/255$ & $\ImageNetFinalTwo{}\pm\ImageNetStdTwo$ &  $\ImageNetBestTwo{}$ &  $\ImageNetDiffTwo{}$ \\
 \bottomrule
\end{tabular}
\end{sc}
\end{small}
\end{center}
\end{table}

\subsection{Robust overfitting: a general phenomenon for adversarially robust deep learning}
In the standard, non-robust deep learning setting, it is common practice to train for as long as possible to minimize the training loss, as modern convergence curves for deep learning generally observe that the testing loss continues to decrease with the training loss. On the contrary, for the setting of adversarially robust training we make the following discovery:

\emph{Unlike the standard setting of deep networks, overfitting for adversarially robust training can result in worse test set performance.}

This phenomenon, which we refer to as ``robust overfitting'', results in convergence curves as shown earlier in Figure \ref{fig:cifar10_overfitting}. Although training appears normal in the earlier stages, after the learning rate decays, the robust test error briefly decreases but begins to increase as training progresses. This behavior indicates that the optimal performance is not obtained at the end of training, unlike in standard training for deep networks.

We find that robust overfitting occurs across a variety of datasets, algorithmic approaches, and perturbation threat models, indicating that it is a general property of the adversarial training formulation and not specific to a particular problem, as can be seen in Table \ref{tab:summary} for $\ell_\infty$ and $\ell_2$ perturbations on SVHN, CIFAR-10, CIFAR-100, and ImageNet. A more detailed and expanded version of this table summarizing the full extent of robust overfitting as well as the corresponding learning curves for each setting can be found in Appendix \ref{app:full_summary}. We consistently find that there is a significant gap between the best robust test performance during training and the final robust test performance at the end of training,
observing an increase of \cifardiff{}\% robust error for CIFAR-10 and \ImageNetDiffInf\% robust error for ImageNet against an $\ell_\infty$ adversary, to highlight a few. Robust overfitting is also not specific to PGD-based adversarial training, and affects faster adversarial training methods such as FGSM adversarial training\footnote{\citet{wong2020fast} also observe a different form of overfitting specifically for FGSM adversarial training which they refer to as ``catastrophic overfitting''. This is separate behavior from the robust overfitting described in this paper, and the specifics of this distinction are discussed further in Appendix \ref{app:catastrophic}.} \citep{wong2020fast} as well as top performing algorithms for adversarially robust training such as TRADES \citep{zhang2019theoretically}.

\begin{figure}[t]
\centering
\includegraphics{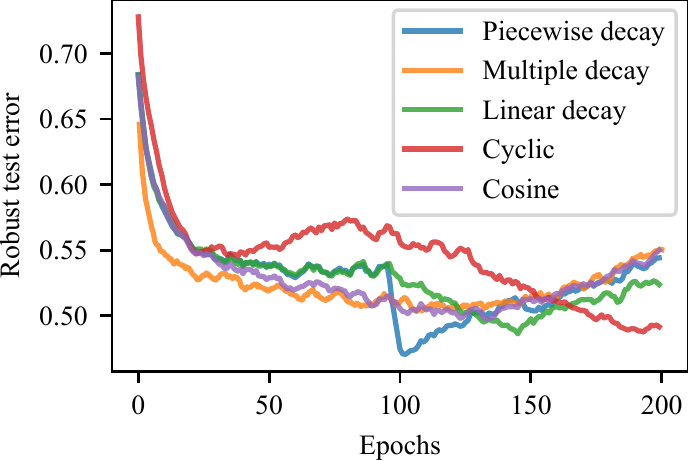}
\caption{Robust test error over training epochs for various learning rate schedules on CIFAR-10. None of the alternative smoother learning rate schedules can achieve a peak performance competitive with the standard piecewise decay learning rate, indicating that the peak performance is obtained by having a single discrete jump. Note that the multiple decay schedule is actually run for 500 epochs, but compressed into this plot for a clear comparison.}
\label{fig:lr_schedules}
\end{figure}

\paragraph{Learning rate schedules and robust overfitting}
Since the change in performance appears to be closely linked with the first drop in the scheduled learning rate decay, we explore how different learning rate schedules affect robust overfitting on CIFAR-10, as shown in Figure \ref{fig:lr_schedules}, with complete descriptions of the various learning rate schedules in Appendix \ref{app:lr-schedules}.
In summary, we find that smoother learning rate schedules (which take smaller decay steps or interpolate the change in learning rate over epochs) simply result in smoother curves that still exhibit robust overfitting. Furthermore, with each smoother learning rate schedule, the best robust test performance during training is strictly worse than the best robust test performance during training with the discrete piecewise decay schedule. In fact, the parameters of the discrete piecewise decay schedule can even be tuned to slightly exacerbate the sudden improvement in performance after the first learning rate decay step, which we discuss further in Appendix \ref{app:lr-tune}

\subsection{Mitigating robust overfitting with early stopping}


\begin{figure}[t]
\centering
\includegraphics{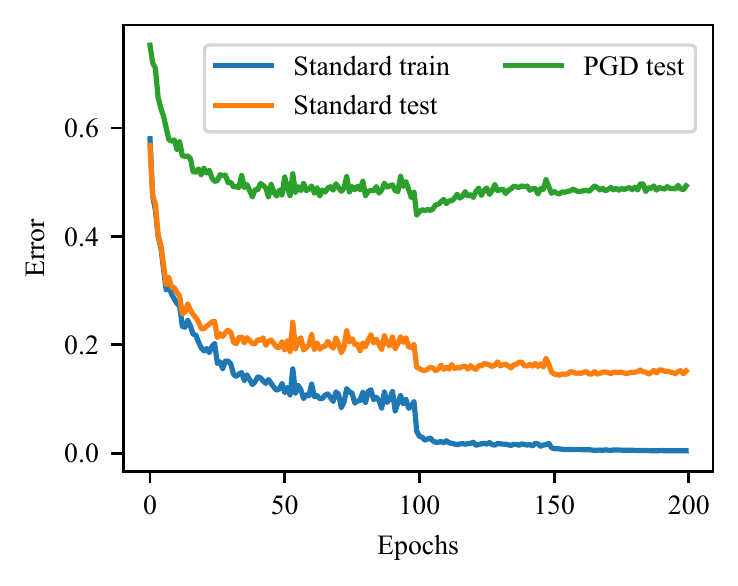}
\caption{Learning curves showing standard and robust error rates for a Wide ResNet model trained with TRADES on CIFAR-10. Early stopping after the initial learning rate decay is crucial in order to achieve the \tradesreported{}\% robust test error reported by \citet{zhang2019theoretically}, which eventually degrades to \tradesfinal{}\% robust test error when the training has converged.}
\label{fig:trades_single}
\end{figure}

Proper early stopping, an old form of implicit regularization, calculates a metric on a hold-out validation set to determine when to stop training in order to prevent overfitting.
Since the test performance does not monotonically improve during adversarially robust training due to robust overfitting, it is advantageous for robust networks to use early stopping to achieve the best possible robust performance.

We find that, for example, the TRADES approach relies heavily on using the best robust performance on the test set from an earlier checkpoint in order to achieve their top reported result of \tradesreported{}\% robust error against an $\ell_\infty$ PGD adversary with radius $8/255$ on CIFAR-10, a number which is typically viewed as a substantial algorithmic improvement in adversarial robustness over standard PGD-based adversarial training.
In our own reproduction of the TRADES experiment, we confirm that allowing the TRADES algorithm to train until convergence results in significant degradation of robust performance as seen in Figure \ref{fig:trades_single}. Specifically, the robust test error of the model at the checkpoint with the best performance on the test set is \tradesbesttrades{}\% whereas the robust test error of the model at the end of training has increased to \tradesfinal{}\%.\footnote{We used the public implementation of TRADES available at \url{https://github.com/yaodongyu/TRADES} and simply ran it to completion using the same learning rate decay schedule used by \citet{madry2017towards}.}

Surprisingly, when we early stop vanilla PGD-based adversarial training, selecting the model checkpoint with the best performance on the test set, we find that PGD-based adversarial training performs just as well as more recent algorithmic approaches such as TRADES. Specifically, when using the \emph{same} architecture (a Wide ResNet with depth 28 and width factor 10)
and the \emph{same} 20-step PGD adversary for evaluation used by  \citet{zhang2019theoretically} for TRADES, the model checkpoint with the best performance on the test set from vanilla PGD-based adversarial training achieves \cifarbesttrades{}\% robust test error, which is actually slightly better than the best reported result for TRADES from \citet{zhang2019theoretically}.\footnote{We found that our implementation of the PGD adversary to be slightly more effective, increasing the robust test error of the TRADES model and the PGD trained model to \tradesbest{}\% and \cifarbest{}\% respectively.}

Similarly, we find early stopping to be a factor in the robust test performance for publicly released pre-trained ImageNet models \citep{robustness}.
Continuing to train these models degrades the robust test performance from \ImageNetBestInf\% to \ImageNetFinalInf\% robust test error for $\ell_\infty$ robustness at $\epsilon=4/255$ and \ImageNetBestTwo\% to \ImageNetFinalTwo\% robust test error for $\ell_2$ robustness at $\epsilon=128/255$. This shows that these models are also susceptible to robust overfitting and benefit greatly from early stopping.\footnote{We use the publicly available framework from \url{https://github.com/madrylab/robustness} and continue training checkpoints obtained from the authors using the same learning parameters.} The corresponding learning curves are shown in Appendix \ref{app:imagenet}.

\begin{figure}[t]
\centering
\includegraphics{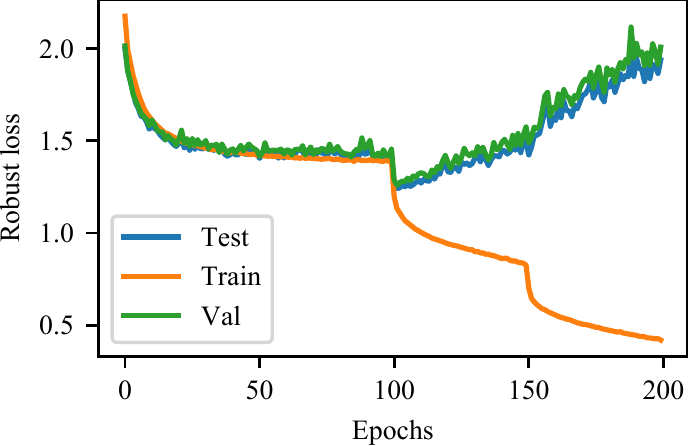}
\caption{Learning curves for a CIFAR-10 pre-activation ResNet18 model trained with a hold-out validation set of 1,000 examples. We find that the hold-out validation set is enough to reflect the test set performance, and stopping based on the validation set is able to prevent overfitting and recover \cifarpreactvalidation{}\% robust test error, in comparison to \cifarpreactbest{}\% achieved by the best-performing model checkpoint.}
\label{fig:validation}
\end{figure}


\paragraph{Validation-based early stopping} Early stopping based on the test set performance, however, leaks test set information and goes against the traditional machine learning paradigm. Instead, we find that it is still possible to recover the best test performance achieved during training with a true hold-out validation set. By holding out 1,000 examples from the CIFAR-10 training set for validation purposes, we use validation-based early stopping to achieve \cifarpreactvalidation{}\% robust error on the test set \emph{without looking at the test set}, in comparison to the \cifarpreactbest{}\% robust error achieved by the best-performing model checkpoint for a pre-activation ResNet18.
The resulting validation curve during training closely matches the testing curve as seen in Figure \ref{fig:validation}, and suggests that although robust overfitting degrades the robust test set performance, selecting the best checkpoint in adversarially robust training for deep networks still does not appear to significantly overfit to the test set (which has been previously observed in the standard, non-robust setting \cite{recht2018cifar}).

\subsection{Reconciling double descent curves}
Modern generalization curves for deep learning typically show improved test set performance for increased model complexity beyond data point interpolation in what is known as \emph{double descent} \citep{belkin2019reconciling}. This suggests that overfitting by increasing model complexity using overparameterized neural networks is beneficial and improves test set performance. However, this appears to be at odds with the main findings of this paper; since training for longer can also be viewed as increasing model complexity, the fact that training for longer results in worst test set performance seems to contradict double descent. 


\begin{figure}[t]
\centering
\includegraphics{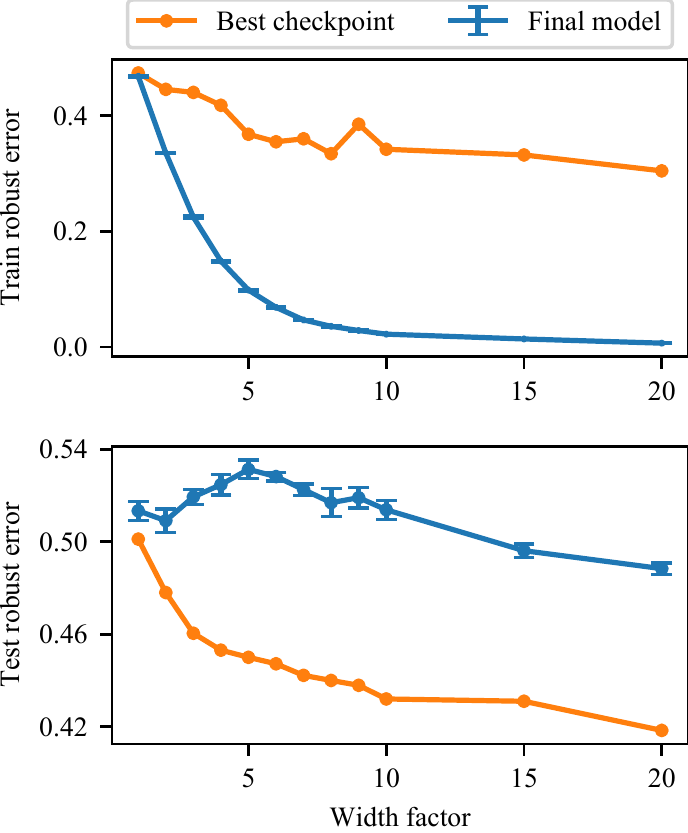}
\caption{Generalization curves depicting double descent for adversarially robust generalization, where hypothesis class complexity is controlled by varying the width factor for a wide residual network. Each final model point represents the average performance over the last 5 epochs with the corresponding width factor from training until convergence. The best checkpoint refers to the lowest robust test error achieved by a model checkpoint during training, and illustrates the significant gap in performance between the best and final models resulting from robust overfitting. }
\label{fig:double_descent}
\vspace{-0.15in}
\end{figure}

We find that, while increasing either training time or architecture size can be viewed as increasing model complexity, these two approaches actually have separate effects; training for longer degrades the robust test set performance regardless of architecture size, while increasing the model architecture size still improves the robust test set performance despite robust overfitting. This was briefly noted by \citet{nakkiran2019deep} for the $\ell_2$ robust setting, and so in this section we show that this generally holds also in the $\ell_\infty$ robust setting. We explore these properties by training multiple adversarially robust Wide ResNets \citep{zagoruyko2016wide} with varying widths to control model complexity. In Figure \ref{fig:double_descent}, we see that no matter how large the model architecture is, robust overfitting still results in a significant gap between the best and final robust test performance. However, we also see that adversarially robust training still produces the double descent generalization curve, as the robust test performance increases and then decreases again with architecture size,
suggesting that the double descent and robust overfitting are separate phenomenon.
Even the lowest robust test error achieved during training continues to descend with increased model complexity, suggesting that larger architecture sizes are still beneficial for adversarially robust training despite robust overfitting. More details and learning curves for a wide range of architecture sizes can be found in Appendix \ref{app:arch_sizes}.

\setlength{\tabcolsep}{6pt}
\begin{table}[t]
\caption{Robust performance of PGD-based adversarial training with different regularization methods on CIFAR-10 using a PreActResNet18 for $\ell_\infty$ with radius $8/255$. The ``best'' robust test error is the lowest test error achieved during training whereas the final robust test error is averaged over the last five epochs. Each of the regularization methods listed is trained using the optimally chosen hyperparameter. Pure early stopping is done with a validation set.}
\begin{center}
\begin{small}
\begin{sc}
\label{tab:reg_compare_table}
\begin{tabular}{llll}
\toprule
& \multicolumn{3}{c}{Robust Test Error ($\%$)} \\
Reg Method & Final & Best & Diff\\
\midrule
 Early stopping w/ val & $\mathbf{\cifarpreactvalidation{}}$ & $\cifarpreactbest{}$ & $0.2$ \\
 $\ell_1$ regularization & $\cifarlonefinal{}\pm \cifarlonestd{}$ & $\cifarlonebest{}$ & $\cifarlonediff{}$ \\
 $\ell_2$ regularization & $\cifarltwofinal{}\pm \cifarltwostd{}$ & $\cifarltwobest{}$ & $\cifarltwofinal{}$ \\
 Cutout & $\cifarcutoutfinal{}\pm \cifarcutoutstd{}$ & $\cifarcutoutbest{}$ & $\cifarcutoutdiff{}$ \\
 Mixup & $\cifarmixupfinal{}\pm \cifarmixupstd{}$ & $\cifarmixupbest{}$ & $\cifarmixupdiff{}$\\
 Semi-supervised & $47.1\pm 4.32$ & $40.2$ & $6.9$ \\
 \bottomrule
\end{tabular}
\end{sc}
\end{small}
\end{center}
\end{table}


\section{Alternative methods to prevent robust overfitting}
\label{sec:exploring}
In this section, we explore whether common methods for combating overfitting in standard training are successful at mitigating robust overfitting in adversarial training. We run a series of ablation studies on CIFAR-10 using classical and modern regularization techniques, yet ultimately find that no technique performs as well in isolation as early stopping, as shown in Table \ref{tab:reg_compare_table} (a more detailed table including standard error can be found in Appendix \ref{app:reg_compare_full_table}). Unless otherwise stated, we begin each experiment with the standard setup for $\ell_\infty$ PGD-based adversarial training with a 10-step adversary with step size $2/255$ using a pre-activation ResNet18 \citep{he2016identity} (details for the training procedure and the PGD adversary can be found in Appendix \ref{app:experiment_setup}). All experiments in this section were run with one GeForce RTX 2080ti unless a Wide ResNet was trained, in which case two GPUs were used.


\begin{figure}[t]
\centering
\includegraphics{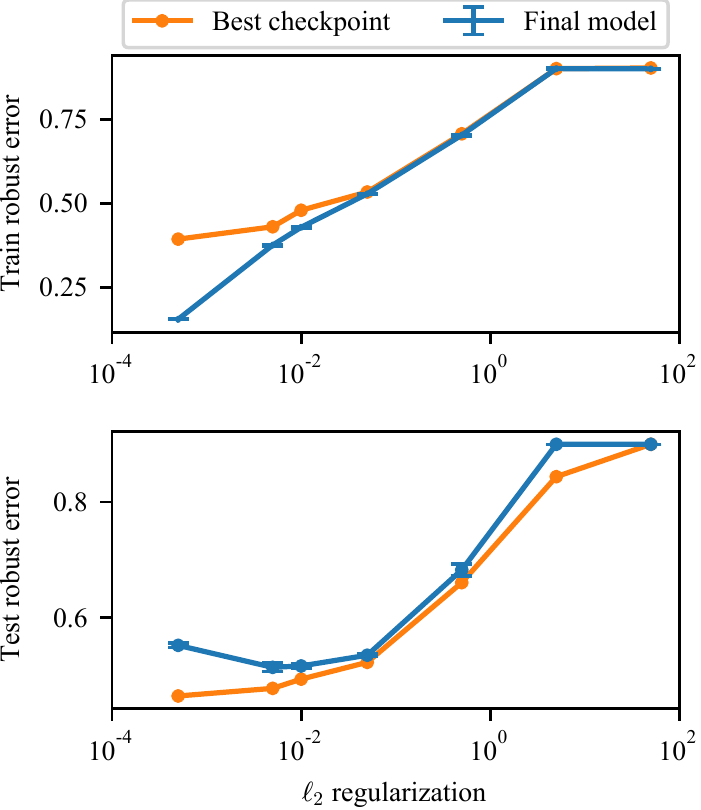}
\caption{Robust performance on the train and test set for varying degrees of $\ell_2$ regularization. $\ell_2$ regularization is unable to match the same performance of early stopping without also using early stopping, even with an optimally chosen hyperparameter of $\lambda = 5 \cdot 10^{-3}$ which achieves \cifarltwofinal{}\% robust test error.}
\label{fig:l2_reg}
\end{figure}

\subsection{Explicit regularization}
A classical method for preventing overfitting is to add an explicit regularization term to the loss, penalizing the complexity of the model parameters. Specifically, the term is typically of the form $\lambda \Omega(\theta)$, where $\theta$ contains the model parameters, $\Omega(\theta)$ is some regularization penalty, and $\lambda$ is a hyperparameter to control the regularization effect. A typical choice for $\Omega$ is $\ell_p$ regularization for $p \in \{1,2\}$, where $\ell_2$ regularization is canonically known as weight decay and commonly used in standard training of deep networks, and $\ell_1$ regularization is known to induce sparsity properties.

We explore the effects of using $\ell_1$ and $\ell_2$ regularization when training robust networks on robust overfitting, and sweep across a range of hyperparameter values as seen in Figure \ref{fig:l2_reg} for $\ell_2$.\footnote{Proper parameter regularization only applies the penalty to the weights $w$ of the affine transformations at each layer, excluding the bias terms and batch normalization parameters.} Although explicit regularization does improve the performance to some degree, on its own, it is still not as effective as early stopping, with the best explicit regularizer achieving \cifarltwofinal{}\% robust test error with $\ell_2$ regularization and parameter $\lambda = 5\cdot 10^{-2}$.
Additionally, neither of these regularization techniques can completely remove the detrimental effects of robust overfitting without drastically over-regularizing the model, which is shown and discussed further in Appendix \ref{app:explicit_regularization}, along with the corresponding plots for $\ell_1$ regularization.

\begin{figure}[t]
\centering
\includegraphics{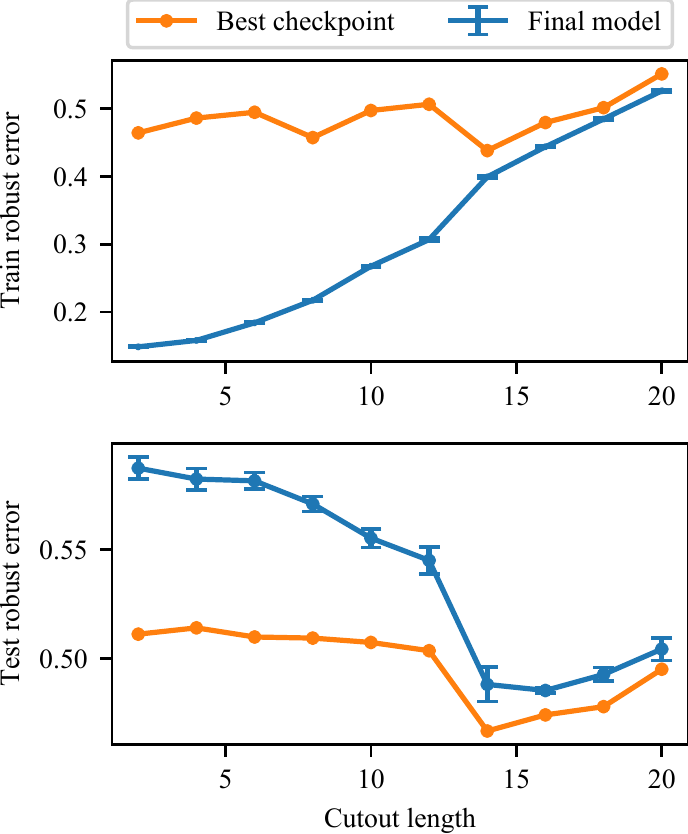}
\caption{Robust performance on the train and test set with cutout across varying patch lengths. Even with the optimal patch length of 14, cutout does not surpass the performance of early stopping, achieving at best \cifarcutoutfinal{}\% robust test error at the end of training.}
\label{fig:cutout}
\end{figure}


\begin{figure}[t]
\centering
\includegraphics{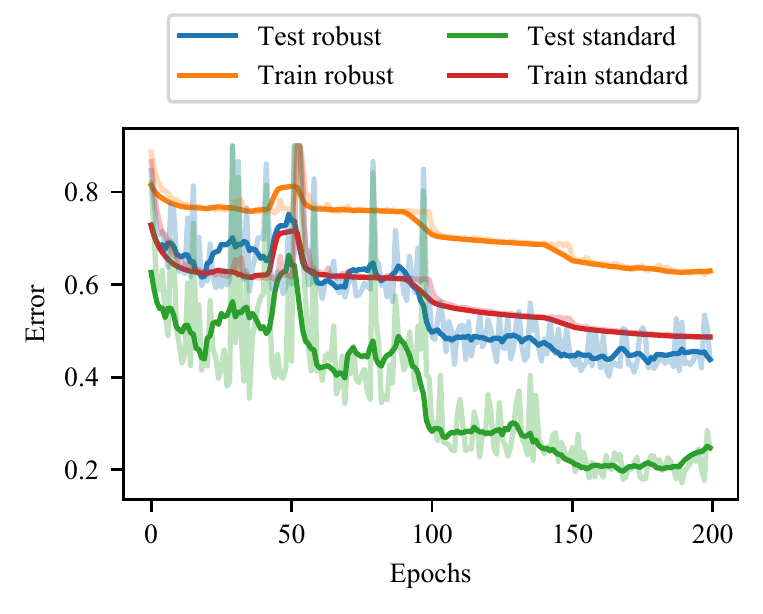}
\caption{Learning curves for robust training with semi-supervised data augmentation, where we do not see a severe case of robust overfitting. When robust training error has converged, there is a significant amount of variance in the robust test error, so the average final model performance is on par with pure early stopping. Combining early stopping with semi-supervised data augmentation to avoid this variance is the only method we find that significantly improves on pure early stopping, reaching \cifarsemibest{}\% robust test error.}
\label{fig:semisupervised}
\end{figure}

\subsection{Data augmentation for deep learning}
Data augmentation has been empirically shown to reduce overfitting in modern deep learning tasks that involve very high-dimensional data by enhancing the quantity and diversity of the training data. Such techniques range from simple augmentations like random cropping and horizontal flipping to more recent approaches leveraging unlabeled data for semi-supervised learning, and some work has argued that robust deep learning requires more data than standard deep learning \citep{schmidt2018adversarially}.

\paragraph{Cutout and mixup}
Recent data augmentations techniques for deep networks, such as cutout \citep{devries2017improved} and mixup \citep{zhang2017mixup}, are known to reduce overfitting and improve generalization in the standard training setting. We scan a range of hyperparameters for these approaches when applicable, and find a similar story to that of explicit $\ell_p$ regularization; either the regularization effect of cutout and mixup is too low to prevent robust overfitting, or too high and the model is over-regularized, as seen in Figures \ref{fig:cutout} for cutout. When trained to convergence, neither cutout nor mixup is as effective as early stopping, achieving at best \cifarcutoutfinal{}\% robust test error for cutout with a patch length of 14 and \cifarmixupfinal{}\% robust test error for mixup with $\alpha=1.4$. \footnote{We used the public implementations of cutout and mixup available at \url{https://github.com/davidcpage/cifar10-fast} and \url{https://github.com/facebookresearch/mixup-cifar10}} The corresponding plots for mixup and the learning curves for both methods are in Appendix \ref{app:data_aug}, where we see significant robust overfitting cutout but less so for mixup, which appears to be more regularized.

\paragraph{Semi-supervised learning}
We additionally consider a semi-supervised data augmentation technique \citep{carmon2019unlabeled, zhai2019adversarially, alayrac2019labels} which uses a standard classifier to label unlabeled data for use in robust training.
Although there is a large gap between best and final robust performance shown in Table \ref{tab:reg_compare_table}, we find that this is primarily driven by high variance in the robust test error during training rather than from robust overfitting, even when the model has converged as seen in Figure \ref{fig:semisupervised}.
Due to this variance, the final model's average robust performance of \cifarsemifinal{}\% robust test error is similar to the performance obtained by early stopping. By combining early stopping with semi-supervised data augmentation, this variance can be avoided. In fact, we find that the combination of early stopping and semi-supervised data augmentation is the only method that results in significant improvement over early stopping alone, resulting in \cifarsemibest{}\% robust test error.
Experimental details and further discussion for this approach can be found in Appendix \ref{app:semisupervised}. \footnote{We used the data from \url{https://github.com/yaircarmon/semisup-adv} containing 500K pseudo-labeled TinyImages}

\section{Conclusion}
Unlike in standard training, overfitting in robust adversarial training decays test set performance during training in a wide variety of settings. While overfitting with larger architecture sizes results in better test set generalization, it does not reduce the effect of robust overfitting. Our extensive suite of experiments testing the effect of implicit and explicit regularization methods on preventing overfitting found that most of these techniques tend to over-regularize the model or do not prevent robust overfitting, and all of them in isolation do not improve upon early stopping.

Especially due to the prevalence of robust overfitting in adversarial training, we particularly urge the community to use validation sets when performing model selection in this regime, and to analyze the learning curves of their models.
This work exposes a key difference in generalization properties between standard and robust training, which is not fully explained by either classic statistics or modern deep learning, and re-establishes the competitiveness of the simplest adversarial training baseline.



\bibliography{robust_overfitting}
\bibliographystyle{icml2020}


\setlength{\tabcolsep}{5.5pt}
\begin{table*}[!ht]
\caption{Performance of adversarially robust training over a variety of datasets, adversarial training algorithms, and perturbation threat models, where the best error refers to the lowest robust test error achieved during training and the final error is an average of the robust test error over the last 5 epochs. We observe robust overfitting to occur across all experiments.}
\begin{center}
\begin{small}
\begin{sc}
\label{tab:full_summary}
\begin{tabular}{lllrrrrrrr}
\toprule
& & & & \multicolumn{3}{c}{Robust Test Error ($\%$)} & \multicolumn{3}{c}{Standard Test Error ($\%$)} \\
Dataset & Adversary & Norm & Radius & \multicolumn{1}{c}{Final} & \multicolumn{1}{c}{Best} & \multicolumn{1}{c}{Diff} & \multicolumn{1}{c}{Final} & \multicolumn{1}{c}{Best} & \multicolumn{1}{c}{Diff}\\
\midrule
\multirow{2}{*}{SVHN} & \multirow{2}{*}{PGD} & $\ell_\infty$ & $8/255$ & $45.6 \pm 0.40$ & $39.0$ & $6.6$ & $10.0 \pm 0.15$ & $10.2$ & $-0.2$\\
 & & $\ell_2$ & $128/255$ & $\svhntwofinal{}\pm \svhntwostd{}$ & $\svhntwobest{}$ & $\svhntwodiff{}$ & $7.0\pm 0.23$ & $7.2$ & $-0.2$\\
 \midrule
 \multirow{6}{*}{CIFAR-10} & \multirow{2}{*}{PGD} & $\ell_\infty$ & $8/255$ & $\cifarfinal{} \pm \cifarfinalstd{}$ & $\cifarbest{}$ & $8.2$ & $13.4 \pm 0.19$ & $13.9$ & $-0.5$ \\
 & & $\ell_2$ & $128/255$ & $31.1 \pm 0.46$ & $28.4$ & $2.7$ & $11.0\pm 0.08$ & $11.3$ & $-0.3$ \\
 \cmidrule[0.25pt]{2-10}
 & \multirow{2}{*}{FGSM} & $\ell_\infty$ & $8/255$ & $59.8 \pm 0.09$ & $53.7$ & $6.1$ & $12.4 \pm 0.21$ & $13.6$ & $-1.2$ \\
 & & $\ell_2$ & $128/255$ & $31.6 \pm 0.18$ & $29.2$ & $2.4$ & $9.9 \pm 0.16$ & $10.5$ & $-0.6$ \\
 \cmidrule[0.25pt]{2-10}
 & \multirow{2}{*}{TRADES} & $\ell_\infty$ & $8/255$ & $\tradesfinal{} \pm 0.31$ & $\tradesbest{}$ & $\tradesdiff{}$ & $14.97 \pm 0.24$ & $15.9$ & $-0.9$\\
 & & $\ell_2$ & $128/255$ & $\tradestwofinal{}\pm\tradestwostd{}$ & $\tradestwobest{}$  & $\tradestwodiff{}$ &  $33.9\pm 0.95$ & $15.7$ & $18.2$\\
 \midrule
 \multirow{2}{*}{CIFAR-100} & \multirow{2}{*}{PGD}& $\ell_\infty$ & $8/255$ & $78.6 \pm 0.39$ & $71.9$ & $6.7$ & $45.9 \pm 0.23$ & $47.3$ & $-1.4$ \\
  &  & $\ell_2$ & $128/255$ & $62.5 \pm 0.09$ & $56.8$ & $5.7$ & $39.9 \pm 0.22$ & $37.5$ & $2.4$ \\
  \midrule
  \multirow{2}{*}{ImageNet} & \multirow{2}{*}{PGD} & $\ell_\infty$ & $4/255$ & $\ImageNetFinalInf{} \pm \ImageNetStdInf{}$ & \ImageNetBestInf{} & \ImageNetDiffInf{} & $50.5 \pm 14.32$ & $37.0$ & $13.5$\\
  &  & $\ell_2$ & $76/255$ & $\ImageNetFinalTwo{} \pm \ImageNetStdTwo{}$ & $\ImageNetBestTwo{}$ & \ImageNetDiffTwo{} & $63.2 \pm 6.80$ & $40.1$ & $23.1$\\
 \bottomrule
\end{tabular}
\end{sc}
\end{small}
\end{center}
\end{table*}
\newpage
\appendix

\begin{figure}[t]
\centering
\includegraphics{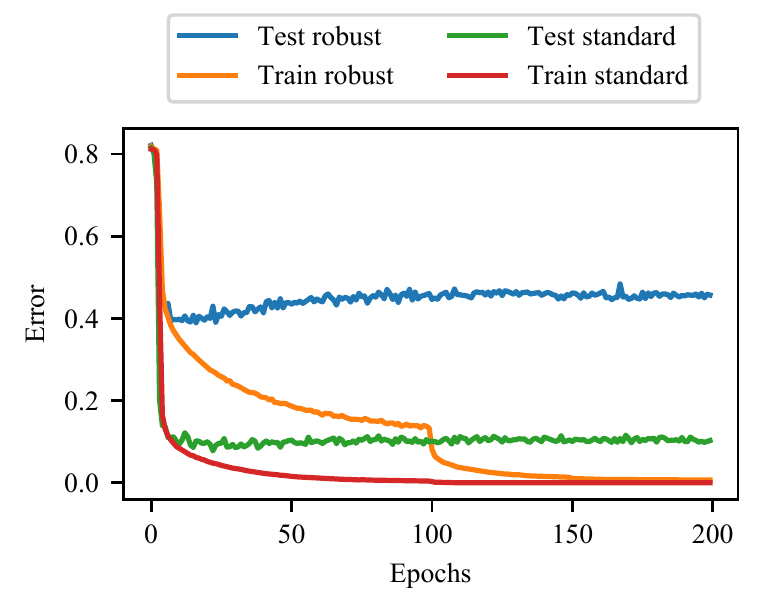}
\caption{Learning curves for training an SVHN classifier which is adversarially robust to $\ell_\infty$ perturbations of radius 8/255. Note that robust overfitting occurs before the learning rate has decayed, likely due to the lower initial learning rate. }
\label{fig:svhn_curve}
\end{figure}

\begin{figure}[t]
\centering
\includegraphics{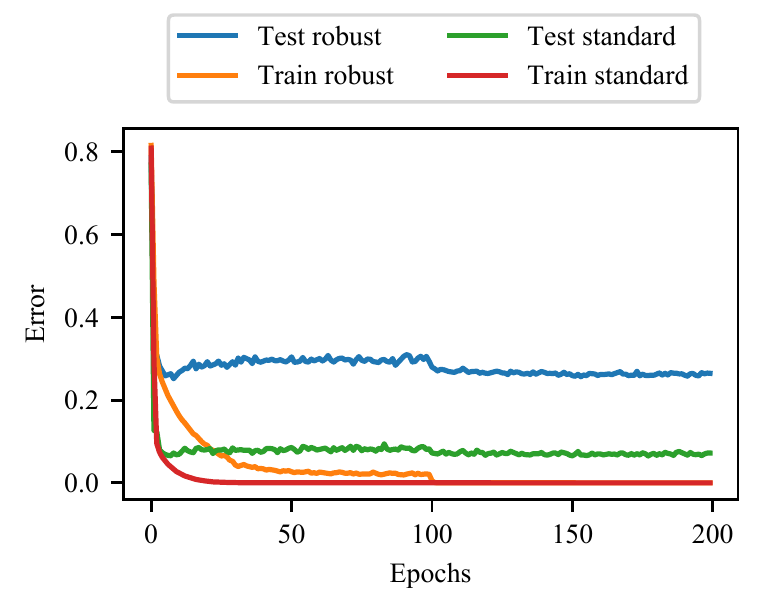}
\caption{Learning curves for training an SVHN classifier which is adversarially robust to $\ell_2$ perturbations of radius 128/255. Robust overfitting occurs early here as well, with robust test error increasing after the 9th epoch. }
\label{fig:svhn_l2_curve}
\end{figure}

\section{Full set of results for Table \ref{tab:summary}}
\label{app:full_summary}
In this section, we extend Table \ref{tab:summary} to additionally include standard error and results from different adversarial training schemes (FGSM and TRADES), as shown in Table \ref{tab:full_summary}. The final error is an average over the final 5 epochs of when the model has converged, along with the standard deviation. The best error is the lowest test error of all model checkpoints during training. For convenience we also show the difference in the final model's error and the best model's error, which indicates the amount of degradation incurred by robust overfitting.

The remainder of this section contains the experimental details for reproducing these experiments, as well as the learning curves for each experiment as visual evidence of robust overfitting. We default to using pre-activation ResNet18s for our experiments, with the exception of Wide ResNets with width factor 10 for $\ell_\infty$ adversaries on CIFAR-10 (for a proper comparison to what is reported for TRADES), and ResNet50s for ImageNet. For CIFAR-10 and CIFAR-100, we train with the SGD optimizer using a batch size of 128, a step-wise learning rate decay set initially at 0.1 and divided by 10 at epochs 100 and 150, and weight decay $5 \cdot 10^{-4}$. For SVHN, we use the same parameters except with a starting learning rate of 0.01 instead. For ImageNet, we use the same learning configuration used to train the pretrained models and simply run them for longer epochs and lower learning rates using the publicly released repository available at \url{https://github.com/madrylab/robustness}.

\paragraph{$\ell_\infty$ adversary} We consider the $\ell_\infty$ threat model with radius 8/255, with the PGD adversary taking 10 steps of size 2/255 on all datasets except for ImageNet. For ImageNet, we fine-tune the pretrained model from \url{https://github.com/madrylab/robustness} \citep{robustness} and continue training with the exact same parameters with a learning rate of 0.001, which uses an adversary with 5 steps of size 0.9/255 within a ball of radius 4/255.

\paragraph{$\ell_2$ adversary} We consider the $\ell_2$ threat model with radius 128/255, with the PGD adversary taking 10 steps of size 15/255 on all datasets except for ImageNet. For Imagenet, we fine-tune the pretrained model from \url{https://github.com/madrylab/robustness} \citep{robustness} and continue training with the exact same parameters with a learning rate of 0.001, which uses an adversary with 7 steps of size 0.5 within a ball of radius 3.




%
%

\begin{figure}[t]
\centering
\includegraphics{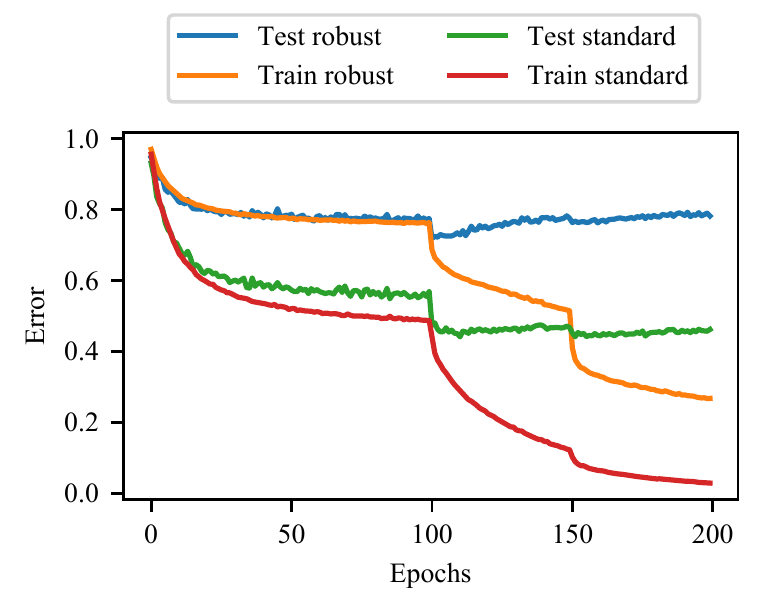}
\caption{Learning curves showing robust overfitting on CIFAR-100 for the $\ell_\infty$ perturbation model.}
\label{fig:cifar100_curve}
\end{figure}

\begin{figure}[t]
\centering
\includegraphics{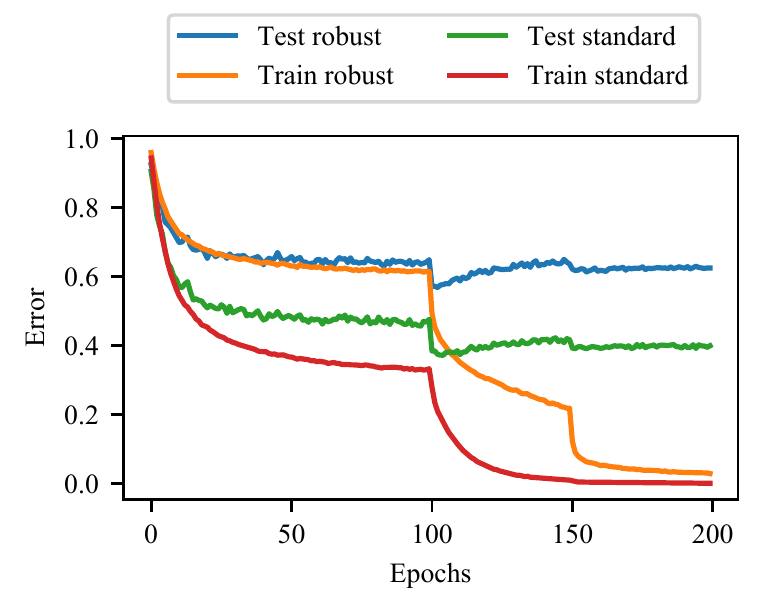}
\caption{Learning curves showing robust overfitting on CIFAR-100 for the $\ell_2$ perturbation model.}
\label{fig:cifar100_l2_curve}
\end{figure}

\subsection{SVHN experiments}
Figures \ref{fig:svhn_curve} and \ref{fig:svhn_l2_curve} contain the convergence plots for the PGD-based adversarial training experiments on SVHN for $\ell_\infty$ and $\ell_2$ perturbations respectively. We find that robust overfitting occurs even earlier on this dataset, before the initial learning rate decay, indicating that the learning rate threshold at which robust overfitting begins to occur has already been passed. The best checkpoint for $\ell_\infty$ achieves 39.0\% robust error, which is a 6.6\% improvement over the 45.6\% robust error achieved at the end of training.

\subsection{CIFAR-100 experiments}
Figures \ref{fig:cifar100_curve} and \ref{fig:cifar100_l2_curve} contain the convergence plots for the PGD-based adversarial training experiments on CIFAR-100 for $\ell_\infty$ and $\ell_2$ perturbations respectively. We find that robust overfitting on this dataset reflects the CIFAR-10 case, occurring after the initial learning rate decay. Note that in this case, both the robust test accuracy and the standard test accuracy are degraded from robust overfitting. The best checkpoint for $\ell_\infty$ achieves 71.9\% robust error, which is a 6.7\% improvement over the 78.6\% robust error achieved at the end of training.

\begin{figure}[t]
\centering
\includegraphics{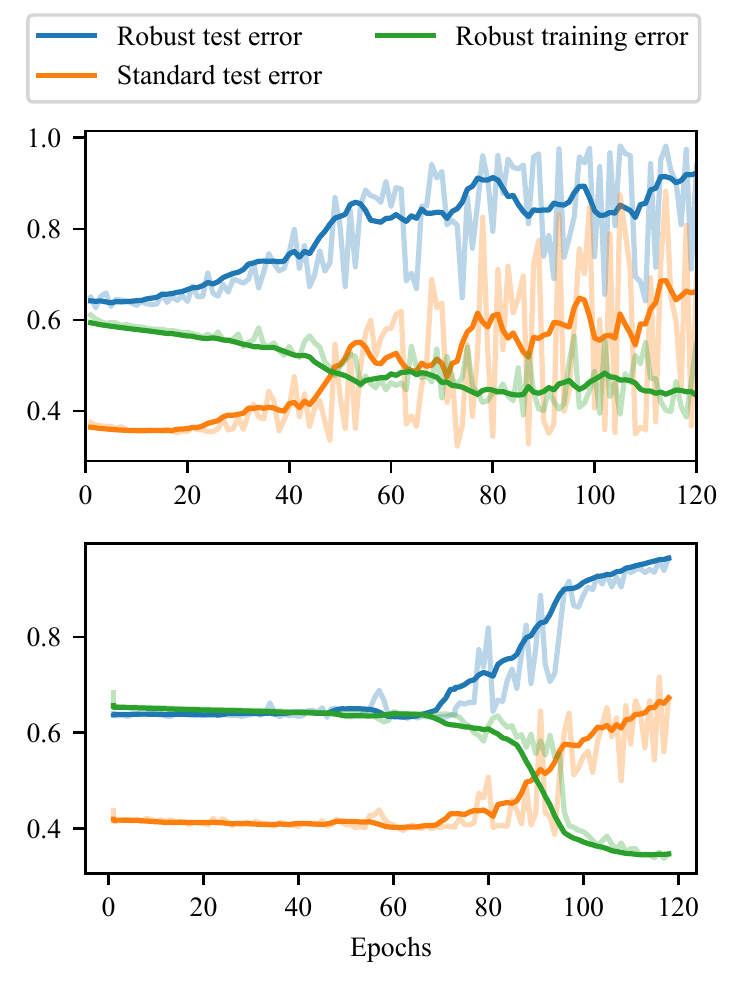}
\caption{Continuation of training released pre-trained ImageNet models for $\ell_\infty$ (top) and $\ell_2$ (bottom). The number of epochs indicate the number of additional epochs the pre-trained models were trained for. }
\label{fig:imagenet_curves}
\end{figure}

\subsection{ImageNet experiments}
\label{app:imagenet}
Figure \ref{fig:imagenet_curves} contains the convergence plots for our continuation of PGD-based adversarial training experiments on ImageNet for $\ell_\infty$ and $\ell_2$ perturbations respectively. Thanks to logs provided by the authors \citep{robustness}, we know the pretrained $\ell_2$ robust ImageNet model had already been trained for 100 epochs at learning rate 0.1 followed by at least 10 epochs at learning rate 0.01, and so we continue training from there and further decay the learning rate at the 150th epoch to 0.001. Logs could not be found for the pretrained $\ell_\infty$ model, and so it is unclear how long it was trained and under what schedule, however the pretrained model checkpoint indicated that the model had been trained for at least one epochs at learning rate 0.001, so we continue training from this point on.

The $\ell_\infty$ pre-trained model appeared to have not yet converged for the checkpointed learning rate, and so further training without any form of learning rate decay was able to gradually deteriorate the performance of the model. The $\ell_2$ pre-trained model seemed to have already converged at the checkpointed learning rate, and so we do not see any significant changes in performance until after decaying the learning rate down to 0.001.

Note that the learning curves here are smoothed by taking an average over a consecutive 10 epoch window, as the actual curves are quite noisy in comparison to other datasets. This noise is reflected in Table \ref{tab:full_summary}, where ImageNet has the greatest variation in final error rates (both robust and standard). Training the models further can in fact improve the performance of the pretrained model slightly at specific checkpoints (e.g. from 66.4\% initial robust test error down to \ImageNetBestInf{}\% robust test error at the best checkpoint for $\ell_\infty$), however eventually the ImageNet models suffer greatly from robust overfitting, with an average increase of \ImageNetDiffInf{}\% robust error for the $\ell_\infty$ model and \ImageNetDiffTwo{}\% robust error for the $\ell_2$ model.

%

\begin{figure}[t]
\centering
\includegraphics{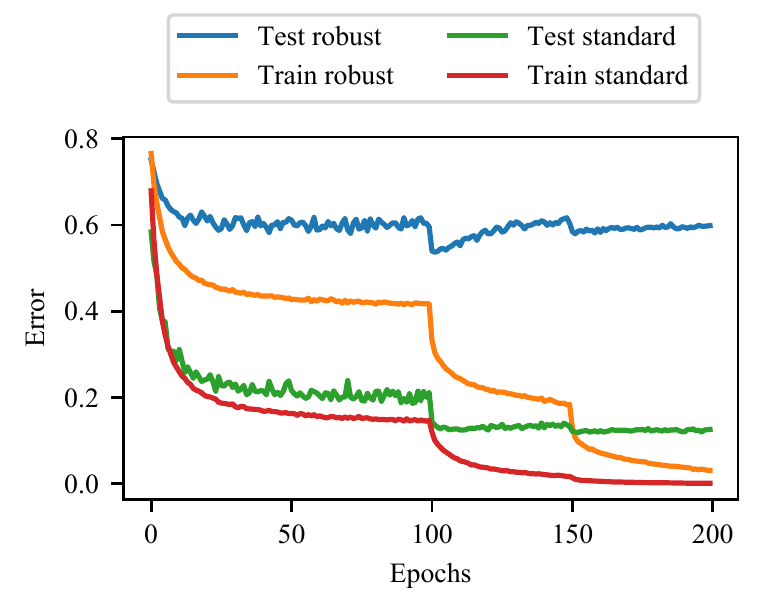}
\caption{Learning curves showing robust overfitting from training with an FGSM adversary on CIFAR-10 for the $\ell_\infty$ perturbation model.}
\label{fig:fgsm_curve}
\vspace{0.1in}
\includegraphics{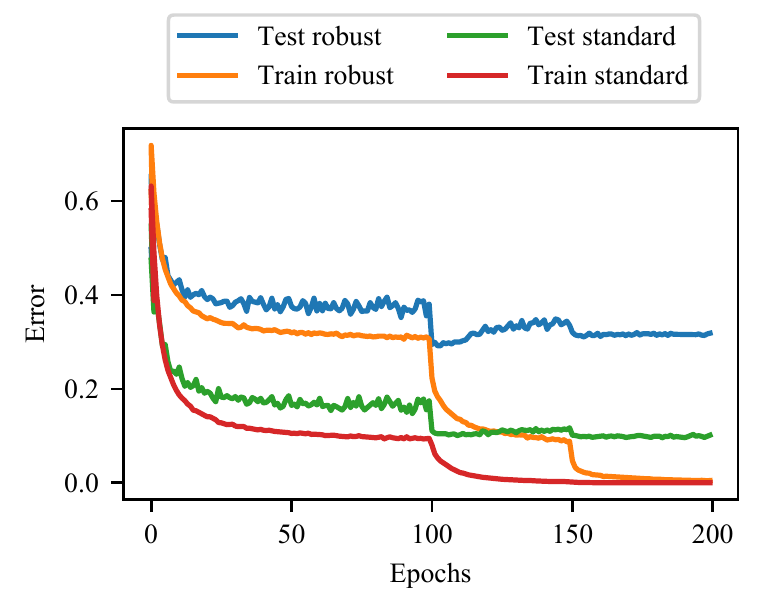}
\caption{Learning curves showing robust overfitting from training with an FGSM adversary on CIFAR-10 for the $\ell_2$ perturbation model.}
\label{fig:fgsm_l2_curve}
\end{figure}

\begin{figure*}[t]
\centering
\includegraphics{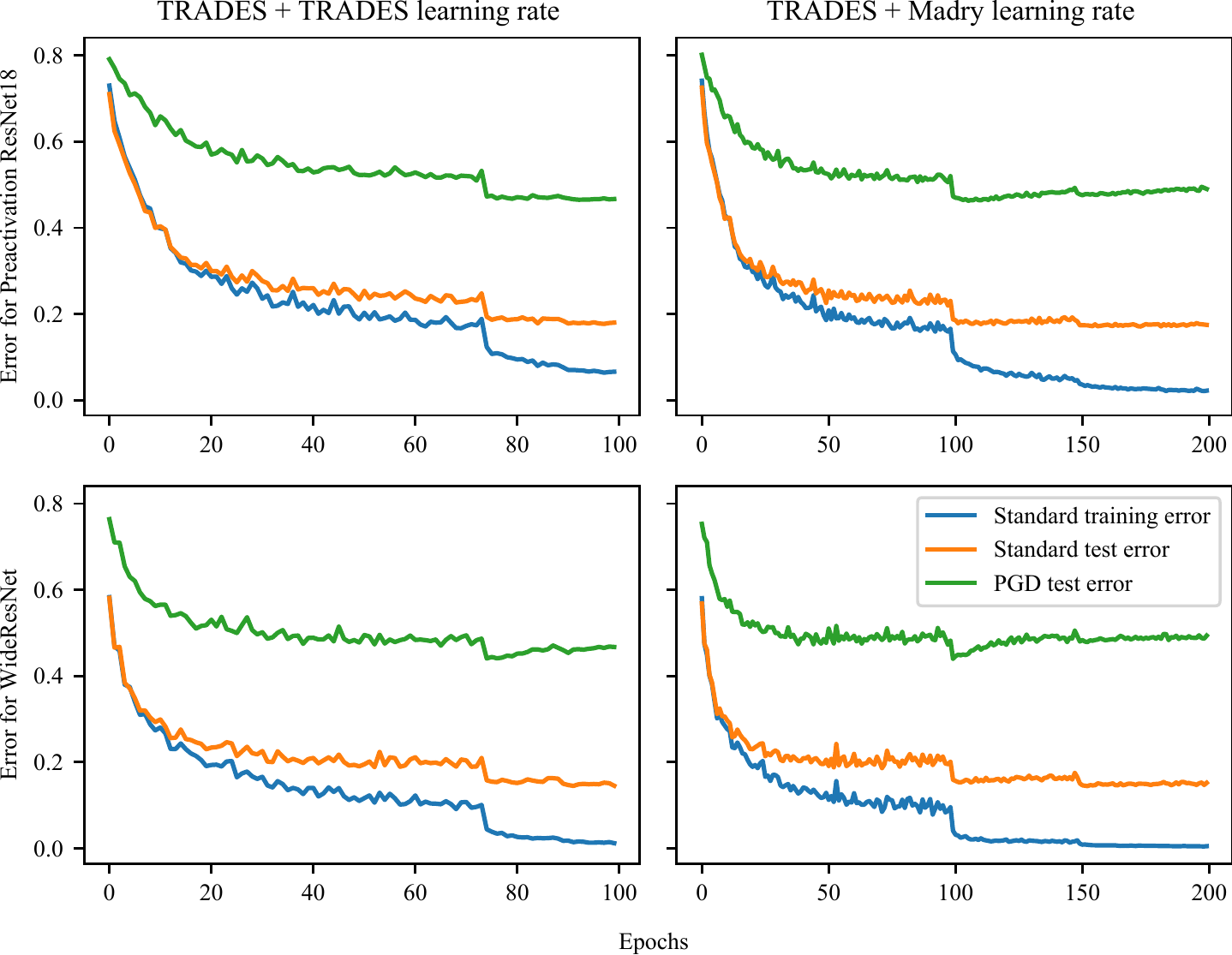}
\caption{Learning curves when running TRADES for robustness to $\ell_\infty$ perturbations of radius 8/255 on combinations of learning rates and architectures for CIFAR10.}
\label{fig:trades_curve}
\end{figure*}

\subsection{CIFAR-10 experiments}
\label{app:cifar10}
\begin{figure}[t]
\centering
\includegraphics{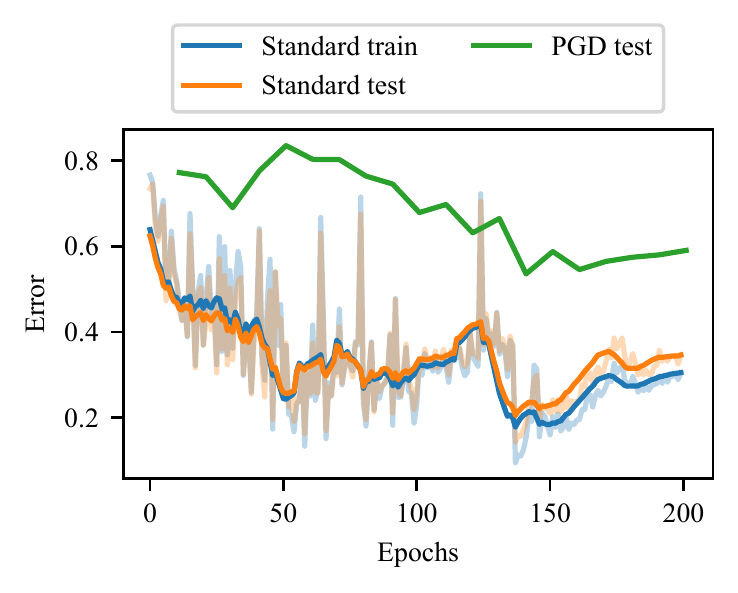}
\caption{Learning curves when running TRADES for robustness to $\ell_2$ perturbations of radius 128/255 for CIFAR10.}
\label{fig:trades_l2_curve}
\end{figure}

For CIFAR-10, in addition to the standard PGD training algorithm, we also consider the FGSM adversarial training algorithm \citep{wong2020fast} and TRADES \citep{zhang2019towards}. The convergence curves showing that robust overfitting still occurs for these two algorithms in both the $\ell_\infty$ and $\ell_2$ setting are shown in Figures \ref{fig:fgsm_curve} and \ref{fig:fgsm_l2_curve} for FGSM and Figures \ref{fig:trades_curve} and \ref{fig:trades_l2_curve} for TRADES.

\label{app:catastrophic}
\paragraph{FGSM adversarial training} For FGSM adversarial training, we use the random initialization described by \citet{wong2020fast}. However, we find that when training until convergence using the piecewise decay learning rate schedule, the recommended step size of $\alpha= $ 10/255 for $\ell_\infty$ training eventually results in catastrophic overfitting. We resort to reducing the step size of the $\ell_\infty$ FGSM adversary to 7/255 to avoid catastrophic overfitting, but still see robust overfitting.

We also note that \citet{wong2020fast} use a cyclic learning rate schedule to further boost the speed of convergence, which differs from the piecewise decay schedule we discuss in this paper. If we run FGSM adversarial training in a more similar fashion to \citet{wong2020fast} with the cyclic learning rate and fewer epochs, we find that this can sidestep the robust overfitting phenomenon and converge directly to the best checkpoint at the end of training. However, this requires a careful selection of the number of epochs: too few epochs and the final model underperforms, whereas too many epochs and we observe robust overfitting. In our setting, we find that training against an FGSM adversary for 50 epochs using a cyclic learning rate with a maximum learning rate of 0.2 allows us to recover a final robust test error of 53.22\%, similar to the best checkpoint of FGSM adversarial training with piecewise decay and 200 epochs which achieved 53.7\% robust test error in Table \ref{tab:full_summary}.



\paragraph{Relation of robust overfitting to catastrophic overfitting} Previous work studying the effectiveness of an FGSM adversary for robust training noted that it is necessary to prevent ``catastrophic overfitting'' in order for FGSM training to be successful, which can be avoided by evaluating a PGD adversary on a training minibatch \citep{wong2020fast}. Here we note that this is a distinct and separate behavior from robust overfitting: while catastrophic overfitting is a product of a model overfitting to a weaker adversary and can be detected by a stronger adversary on the training set, robust overfitting is a degradation of robust test set performance under the \emph{same} adversary used during training which \emph{cannot be detected on the training set}. Indeed, even successful FGSM adversarial training can suffer from robust overfitting when given enough epochs without catastrophically overfitting, as shown in Figure \ref{fig:fgsm_curve}, suggesting that this is related to the generalization properties of adversarially robust training rather than the strength of the adversary.

\paragraph{TRADES} For TRADES we use the publicly released implementation of both the defense and attack available at \url{https://github.com/yaodongyu/TRADES} to remove the potential for any confounding factors resulting from differences in implementation. We consider two possible options for learning rate schedules: the default schedule used by TRADES which decays at 75 and 90 epochs and runs for 100 epochs total (denoted TRADES learning rate),\footnote{This is the learning rate schedule described in the paper by \citet{zhang2019theoretically}. Note that this differs slightly from the implementation in the TRADES repository, which uses the same schedule but only trains for 76 epochs, which is one more epoch after decaying. In our reproduction of the TRADES experiment, the checkpoint after the initial learning rate decay ends up with the best test performance over all 100 epochs.} and the standard learning rate schedule used by \citet{madry2017towards} for PGD adversarial training, which decays at 100 epochs and 150 epochs. We additionally explore both the pre-activation ResNet18 architecture that we use extensively in this paper, as well as the Wide ResNet architecture which TRADES uses. The corresponding learning curves for each combination of learning rate and model can be found in Figure \ref{fig:trades_curve} for $\ell_\infty$.

We note that in three of the four cases, we see a clear instance of robust overfitting. Only the default learning rate schedule used by TRADES on the smaller, pre-activation ResNet18 model doesn't indicate any degradation in robust test set performance. This is likely due the shortened learning rate schedule which implicitly early stops combined with the regularization induced by a smaller architecture having less representational power. The results here are consistent with our earlier findings on the impact of architecture size, where the Wide ResNet architecture achieves better performance than the ResNet18. The shortened TRADES learning rate schedule does not show the full extent of robust overfitting, as the models have not yet converged, whereas the Madry learning rate does (and also achieves a slightly better best checkpoint).

Figure \ref{fig:trades_l2_curve} shows a corresponding curve for $\ell_2$ robustness using TRADES for the pre-activation ResNet18 model with the Madry learning rate, which was the optimal combination from $\ell_\infty$ training. Note that the TRADES repository does not provide default training parameters or a PGD adversary for $\ell_2$ training on CIFAR-10 nor could we find any such description in the corresponding paper, and so we used our attack parameters which were successful for PGD-based adversarial training (10 steps of size 15/255).

\begin{figure*}[]
\centering
\includegraphics{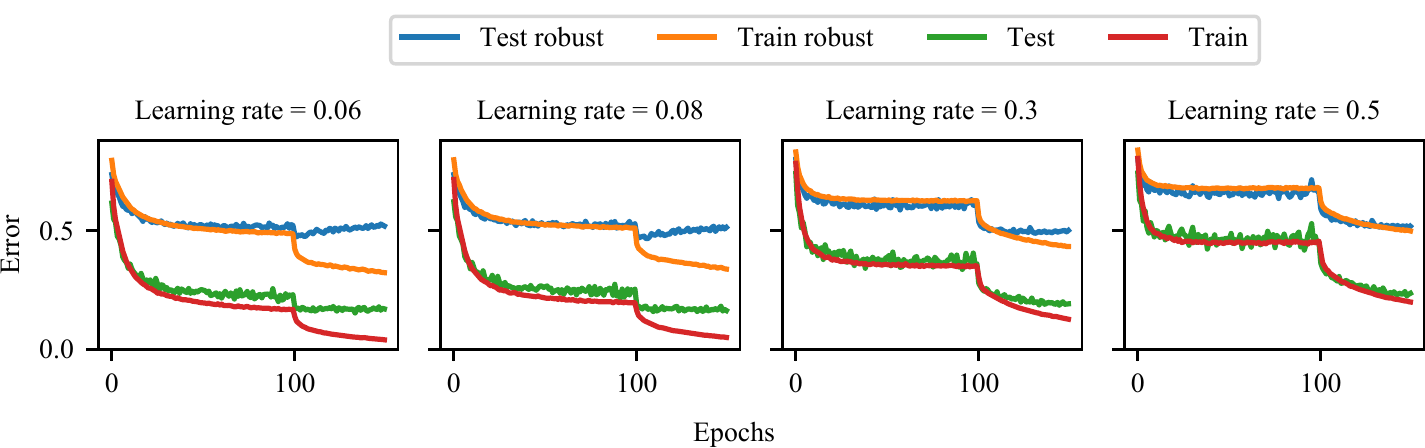}
\caption{Learning curves for a piecewise decay schedule with a modified starting learning rate.}
\label{fig:startlr_curves}
\vspace{0.1in}
\includegraphics{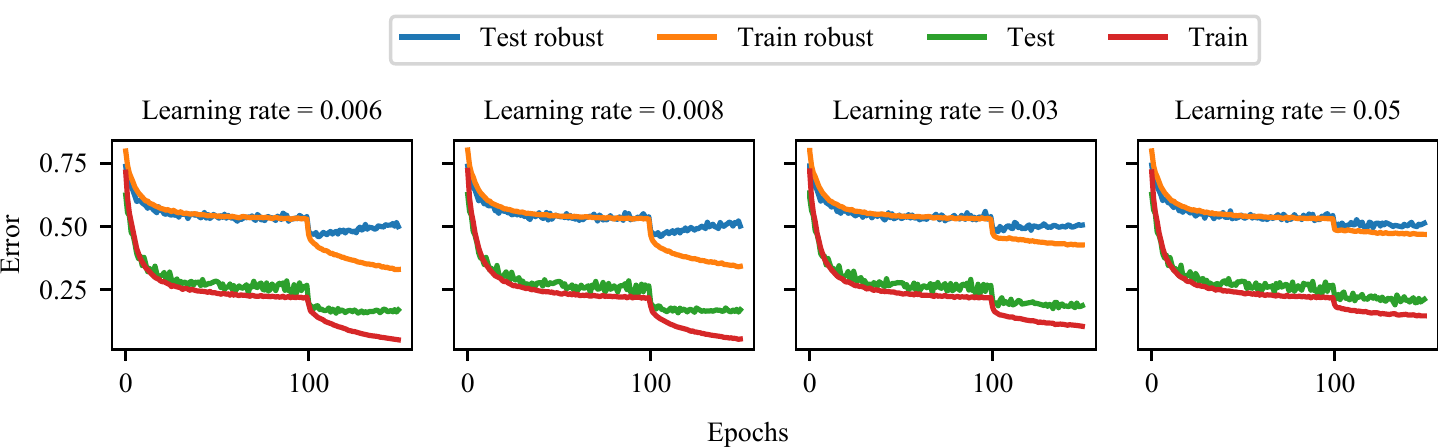}
\caption{Learning curves for a piecewise decay schedule with a modified ending learning rate.}
\label{fig:droplr_curves}
\vspace{0.1in}
\includegraphics{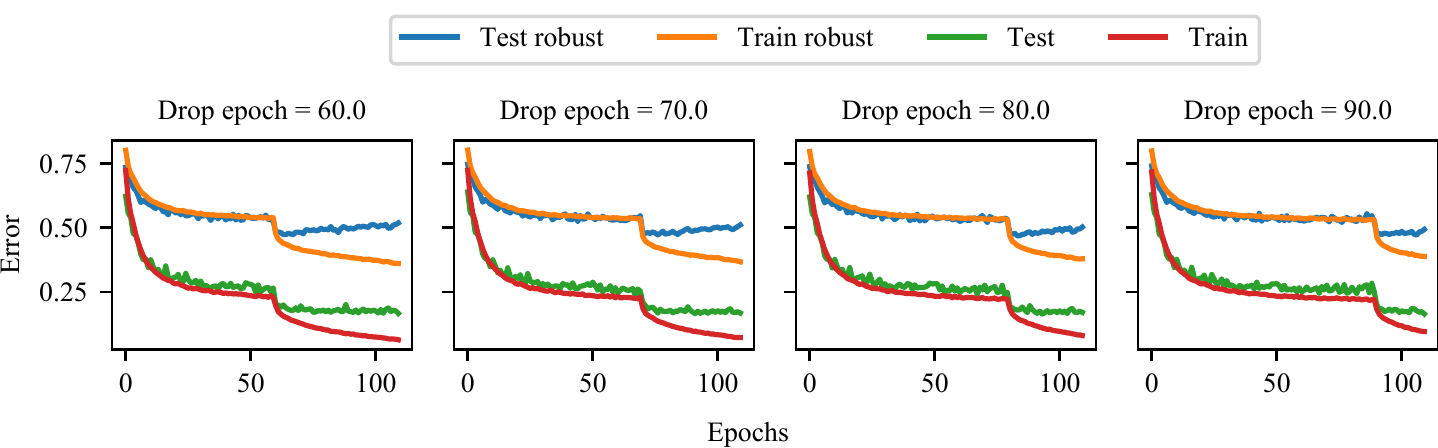}
\caption{Learning curves for a piecewise decay schedule with a modified epoch at which the decay takes effect.}
\label{fig:dropepoch_curves}
\end{figure*}

%

\section{Experiments for various learning rate schedules}
In this section, we explore the effect of the learning rate schedule with greater detail on the CIFAR10 dataset with a pre-activation ResNet18. Our search begins with a sweep over a range of different potential schedules which are commonly used in deep learning. Following this, we tune the best learning rate schedule to investigate its effect on the prevalence of robust overfitting.

\subsection{Different types of schedules}
\label{app:lr-schedules}
We consider the following types of learning rates for our setting.
\begin{enumerate}
\item {\bf Piecewise decay: } This is a fairly common learning rate used in deep learning, which decays the learning rate by a constant factor at fixed epochs. We begin with a learning rate of 0.1 and decay it by a factor of 10 at the 100th and 150th epochs, for 200 total epochs.
\item {\bf Multiple decay: } This is a more gradual version of the piecewise decay schedule, with a piecewise constant schedule which reduces the learning rate at a linear rate in order to make the drop in learning rate less drastic. Specifically, the learning rate begins at 0.1 and is reduced by 0.01 every 50 epochs over 500 total epochs, eventually reaching a learning rate of 0.01 in the last 50 epochs.
\item {\bf Linear decay: } This schedule does a linear interpolation of the drop from 0.1 to 0.01, resulting in a piecewise linear schedule. The learning rate is trained at 0.1 for the first 100 epochs, then linearly reduced down to 0.01 over the next 50 epochs, and further trained at 0.01 for the last 50 epochs for a total of 200 epochs.
\item {\bf Cyclic:} This schedule grows linearly from 0 to to some maximum learning rate $\lambda$, and then is reduced linearly back to 0 over training as proposed by \citet{smith2017cyclical}. We adopt the version from \citet{wong2020fast} which already computed the maximum learning rate for the CIFAR10 setting on the same architecture which peaks 2/5 of the way through training at a learning rate of 0.2 over 200 epochs.
\item {\bf Cosine:} This schedule reduces the learning rate using the cosine function to interpolate from 0.1 to 0 over 200 epochs. This type of schedule was used by \citet{carmon2019unlabeled} when leveraging semi-supervised data augmentation to improve adversarial robustness.
\end{enumerate}
Note that the piecewise decay schedule is the primary learning rate schedule used in this paper. All of these approaches beyond the standard piecewise decay schedule dampen the initial drop in robust test error experienced by the piecewise decay schedule. As a result, the best checkpoints of these alternatives end up with worse performance than the best checkpoint of the piecewise decay schedule, since all of the learning rates eventually start increasing in robust test error due to robust overfitting after the initial drop. Robust overfitting appears to be ubiquitous across different schedules, as most approaches achieve their best checkpoint well before training has converged.

The cyclic learning rate is the exception here, which has two phases corresponding to when the learning rate is growing and shrinking, with the best checkpoint occurring near the end of the second phase. In both phases, the robust performance begins to improve, but then robust overfitting eventually occurs and keeps the model from improving any further. We found that stretching the cyclic learning rate over a longer number of epochs (e.g. 300) results in a similar learning curve but with worse robust test error for both the best checkpoint and the final converged model.

\setlength{\tabcolsep}{4.8pt}
\begin{table}[t]
\caption{Tuning experiments using stochastic gradient descent to optimize the best robust test error obtained from the piecewise decay schedule for a pre-activation ResNet18 on CIFAR-10.}
\begin{center}
\begin{small}
\begin{sc}
\label{tab:tuning}
\begin{tabular}{cccc}
\toprule
Decay epoch & Start LR & End LR & Best Rob Err \\
\midrule
$100$ & $0.1$ & $0.01$ & $46.7\%$\\
\midrule
$60$ & \multirow{4}{*}{$0.1$} & \multirow{4}{*}{$0.01$} & $47.4\%$ \\
$70$ &  &  & $47.3\%$ \\
$80$ &  &  & $46.9\%$ \\
$90$ &  &  & $47.3\%$ \\
 \midrule
\multirow{4}{*}{$100$} & $0.06$ & \multirow{4}{*}{$0.01$} & $47.4\%$ \\
 & $0.08$&  & $46.7\%$ \\
 & $0.3$ &  & $48.7\%$ \\
 & $0.5$ &  & $51.0\%$ \\
 \midrule
\multirow{4}{*}{$100$} & \multirow{4}{*}{$0.1$} & $0.006$ & {\bf $46.0\%$} \\
 &  & $0.008$ & $46.1\%$ \\
 &  & $0.03$ & $47.8\%$ \\
 &  & $0.05$ & $49.3\%$ \\
 \bottomrule
\end{tabular}
\end{sc}
\end{small}
\end{center}
\end{table}

\begin{figure*}[]
\begin{center}
\includegraphics{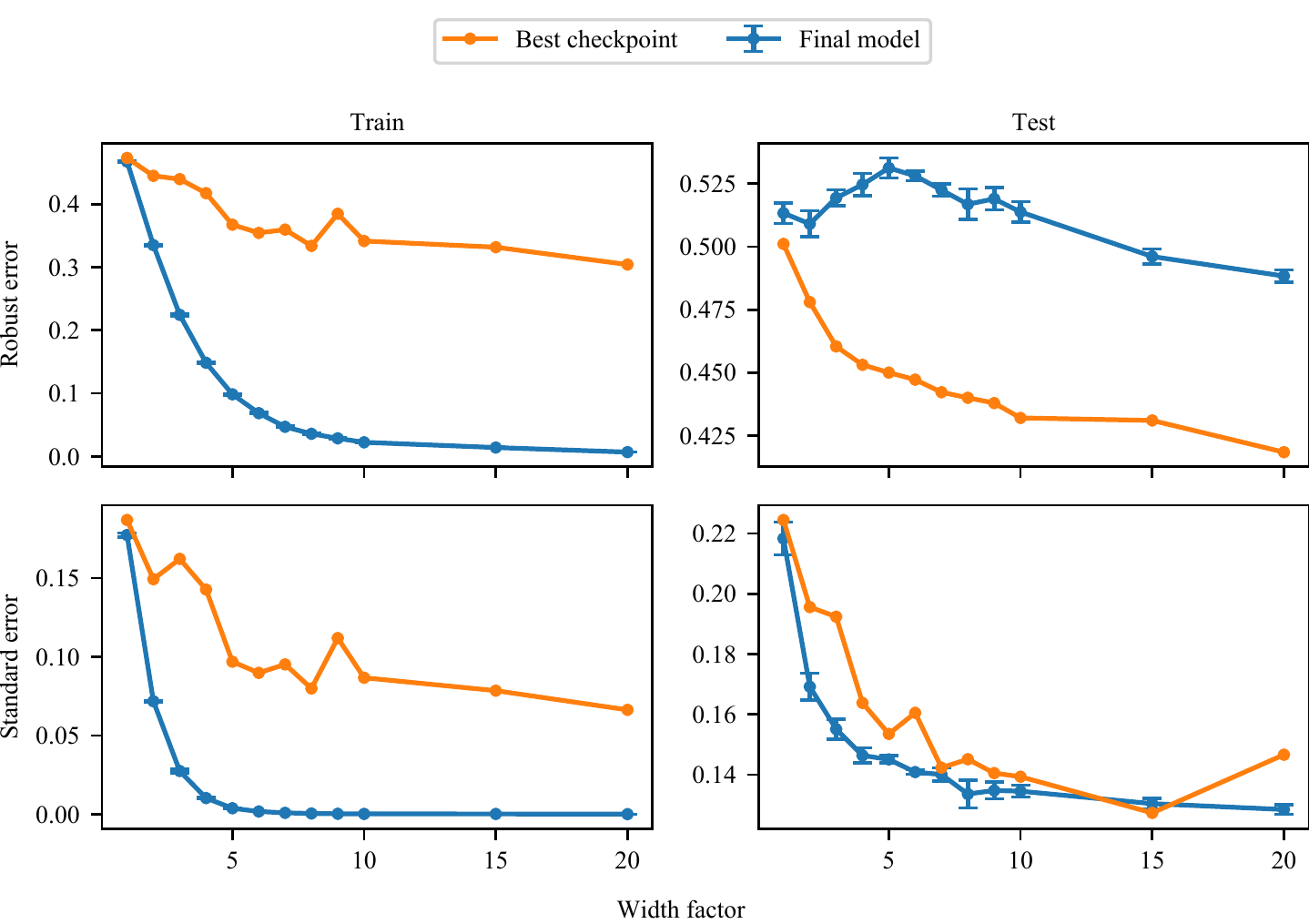}
\caption{Standard and robust performance on the train and test set across Wide ResNets with varying width factors.}
\label{fig:width_generalization_all}

\vspace{0.1in}
\includegraphics{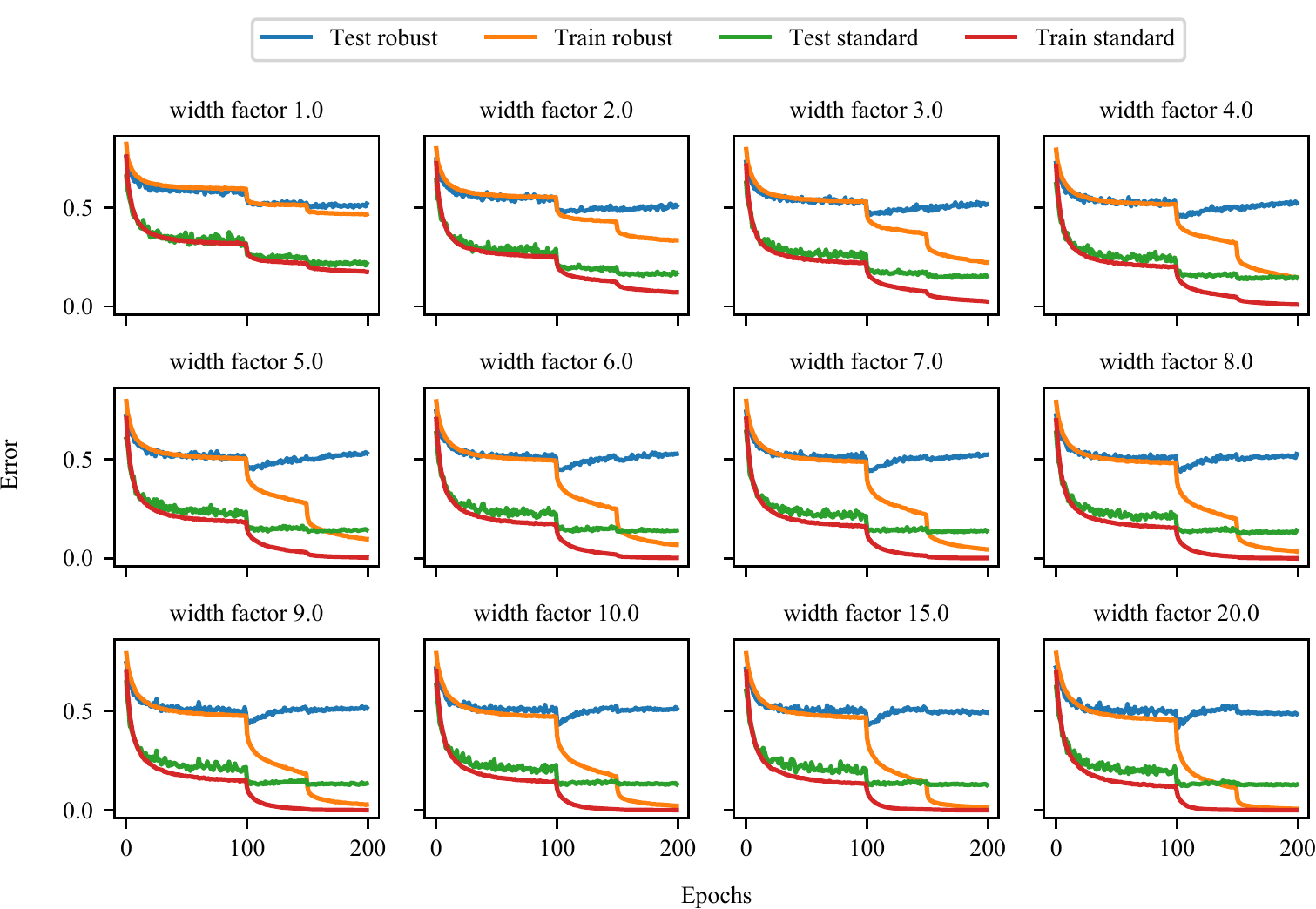}
\caption{Learning curves for training Wide ResNets with different width factors.}
\label{fig:width_curves}
\end{center}
\end{figure*}

\subsection{Tuning the piecewise decay schedule}
\label{app:lr-tune}

Since the piecewise decay schedule appeared to be the most effective method for finding a model with the best robust performance, we investigate whether this schedule can be potentially tuned to improve the robust performance of the best checkpoint even further. The discrete piecewise decay schedule has three possible parameters: the starting learning rate, the ending learning rate, and the epoch at which the decay takes effect. We omit the last 50 epochs of the final decay, since the bulk of the impact from robust overfitting occurs shortly after the first decay in this setting.

While tuning the starting learning rate and the decay epoch largely results in either similar or worse performance, we find that adjusting the learning rate used after the decay epoch can actually slightly improve the robust performance of the best checkpoint by 0.5\%, as seen in Table \ref{tab:tuning}. Note that robust overfitting still occurs in these tuned learning rate schedules as seen in Figures \ref{fig:startlr_curves}, \ref{fig:droplr_curves}, and \ref{fig:dropepoch_curves}, which show the learning curves for each one of the models
shown in Table \ref{tab:tuning}. 

\section{Double descent: exploring architecture sizes}
\label{app:arch_sizes}

For architecture size experiments, we use a Wide ResNet architecture \citep{zagoruyko2016wide} with depth 28 and varying widths to control the size of the network. For each width tested, we plot the standard and robust performance from the best checkpoint and final model in Figure \ref{fig:width_generalization_all}. Learning curves for each width can be found in Figure \ref{fig:width_curves}. All models were trained with the same training parameters described in Section \ref{sec:exploring}. Mean and standard deviation of the final model was taken over the last 5 epochs.

From both the generalization curves and the individual convergence plots, we see that no matter how large the architecture is, the checkpoint which achieves the lowest robust test error always has higher training robust error than the final model at convergence. We also find that both the final model at the end of convergence as well as the best checkpoint found during training all benefit from the increase in architecture size. Consequently, we find that robust overfitting and double descent can occur at the same time, despite having seemingly opposite effects on the notion of overfitting.

In contrast to the standard setting, we observe that the double descent occurs well before robust interpolation of the training data at a width factor of 5, after which the robust test set performance of the final model continues to improve with even larger architecture sizes. The network with width factor 20, the largest that we could run on our hardware, achieves \cifarwidefinal{}\% robust test error at the end of training and \cifarwidebest{}\% robust test error at the best checkpoint. This marks a further improvement over the more typical choice of width factor 10 which achieves \cifarfinal{}\% robust test error at the end of training and \cifarbest{}\% robust test error at the best checkpoint.

\setlength{\tabcolsep}{6pt}
\begin{table*}[ht]
\caption{Performance of adversarially robust training over a variety of regularization techniques for PGD-based adversarial training on CIFAR-10 for $\ell_\infty$ with radius $8/255$.}
\begin{center}
\begin{small}
\begin{sc}
\label{tab:reg_compare_full_table}
\begin{tabular}{lllllll}
\toprule
& \multicolumn{3}{c}{Robust Test Error ($\%$)} & \multicolumn{3}{c}{Standard Test Error ($\%$)} \\
Regularization method & Final & Best & Diff & Final & Best & Diff\\
\midrule
 Early stopping w/ val & $\mathbf{46.9}$ & $46.7$ & $\mathbf{0.2}$ & $18.2$ & $18.2$ & $0.0$\\
 $\ell_1$ regularization & $53.0 \pm 0.39$ & $48.6$ & $4.4$ & $15.9 \pm 0.13$ & $15.4$ & $0.5$ \\
 $\ell_2$ regularizaiton & $51.4 \pm 0.73$ & $46.4$ & $8.8$ & $15.7 \pm 0.21$ & $14.9$ & $0.8$ \\
 Cutout & $48.8 \pm 0.79$ & $46.7$ & $2.1$ & $16.8 \pm 0.21$ & $16.4$ & $0.4$ \\
 Mixup & $49.1 \pm 1.32$ & $46.3$ & $2.8$ & $23.3 \pm 3.04$ & $19.0$ & $4.3$ \\
 Semi-supervised & $47.1$ & $40.2$ & $6.9$ & $23.0 \pm 3.82$ & $17.2$ & $5.8$ \\
 \bottomrule
\end{tabular}
\end{sc}
\end{small}
\end{center}
\end{table*}

\section{Preventing overfitting}
\subsection{Experimental setup}
\label{app:experiment_setup}
For the experiments in preventing overfitting, we use a PGD adversary with random initialization and 10 steps of step size 2/255. This is a slightly stronger adversary than considered in \citet{madry2017towards} by using 3 additional steps, and we found the attack to be more effective than the adversary implemented by TRADES, achieving approximately 1\% more PGD error than the TRADES adversary. However, our goal here is to explore the prevention of robust overfitting, and so it is not necessary to have strongest possible adversarial attack, and so for our purposes this adversary is good enough (and is known to be reasonably strong in the $\ell_\infty$ setting). For training, we use the same parameters as used for the CIFAR-10 experiments in Appendix \ref{app:cifar10} (batch size, learning rate, weight decay, number of epochs). We primarily use the pre-activation ResNet18 since it is already sufficient for exhibiting the robust overfitting behavior.

\subsection{Full set of results for Table \ref{tab:reg_compare_table}}
\label{app:reg_compare_full_table}

In this section, we present the expanded version of Table \ref{tab:reg_compare_table} to include standard test error metrics. The final robust and standard errors are an average of over the final 5 epochs of training when the model has converged, from which the standard deviation is also computed. The one exception is validation-based early stopping, where the final error is taken from the checkpoint chosen by the validation set, and consequently does not have a standard deviation. The best robust error is the lowest test robust error of all checkpoints through training, and the best standard error is the corresponding standard error which comes from this same checkpoint. For convenience we also show the difference in the final model's error and the best model's error, which indicates the amount of degradation incurred by robust overfitting.

\subsection{Explicit regularization}
\label{app:explicit_regularization}
In this section, we extend the plots depicting the robust and standard error over various regularization hyperparameters to also show the performance on the training set. We also show the learning curves for models trained with explicit regularization to show the extent of robust overfitting on various hyperparameter choices.

\begin{figure*}[ht]
\centering
\includegraphics[scale=1]{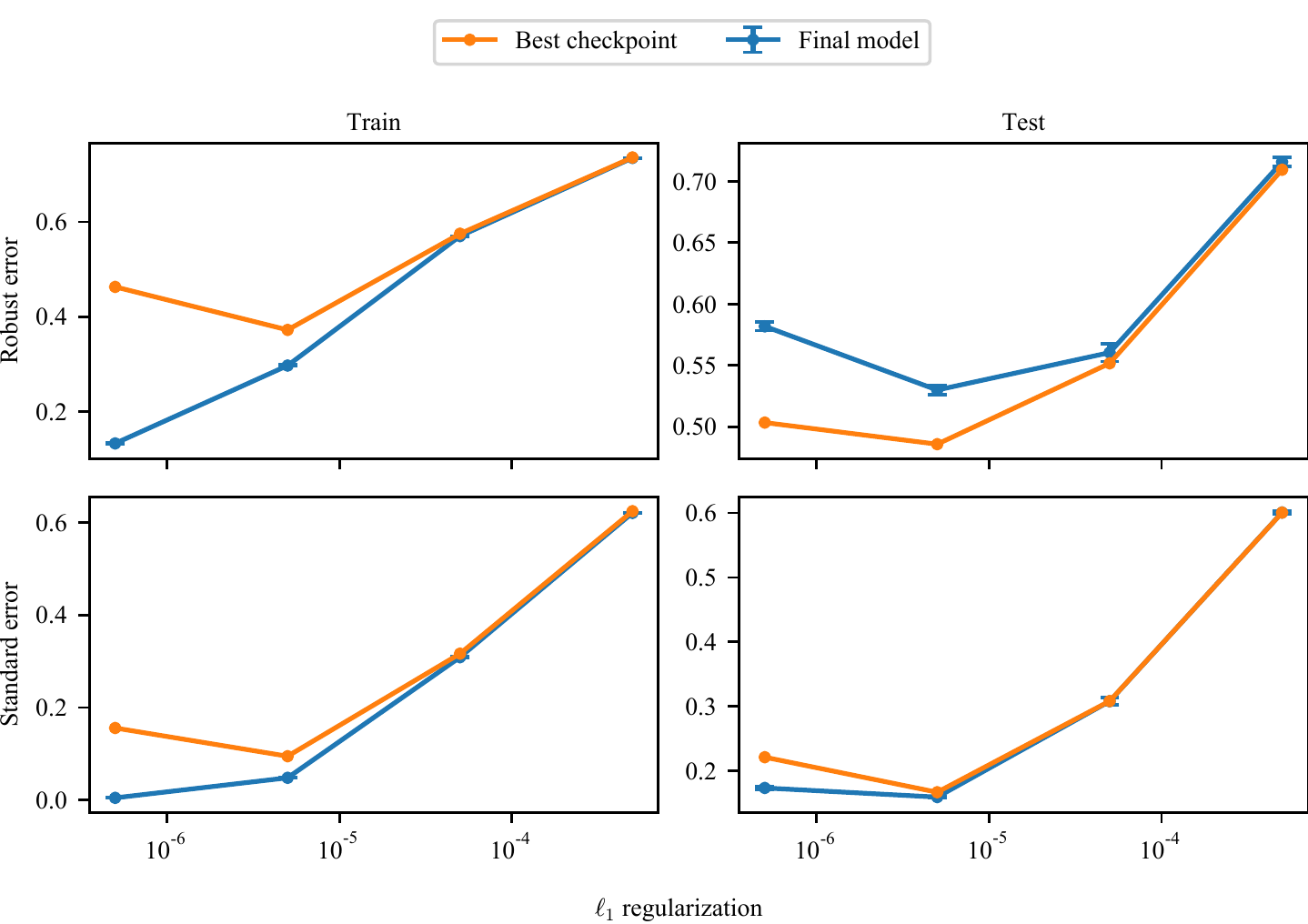}
\caption{Standard and robust performance on the train and test set using varying degrees of $\ell_1$ regularization.}
\label{fig:l1_generalization_all}
\end{figure*}

\paragraph{$\ell_1$ regularization}
Figure \ref{fig:l1_generalization_all} shows the training and testing performance of models using various degrees of $\ell_1$ regularization. We performed a search over regularization parameters $\lambda = \{ 5\cdot 10^{-6}, 5\cdot 10^{-5}, 5\cdot 10^{-4}, 5\cdot 10^{-3}\}$, and found that both the final checkpoint and the best checkpoint have an optimal regularization parameter of $5\cdot 10^{-5}$. Note that we only see robust overfitting at smaller amounts of regularization, since the larger amounts of regularization actually regularize the model to the point where the performance is being severely hurt.

\begin{figure*}[ht]
\centering
\includegraphics[scale=1]{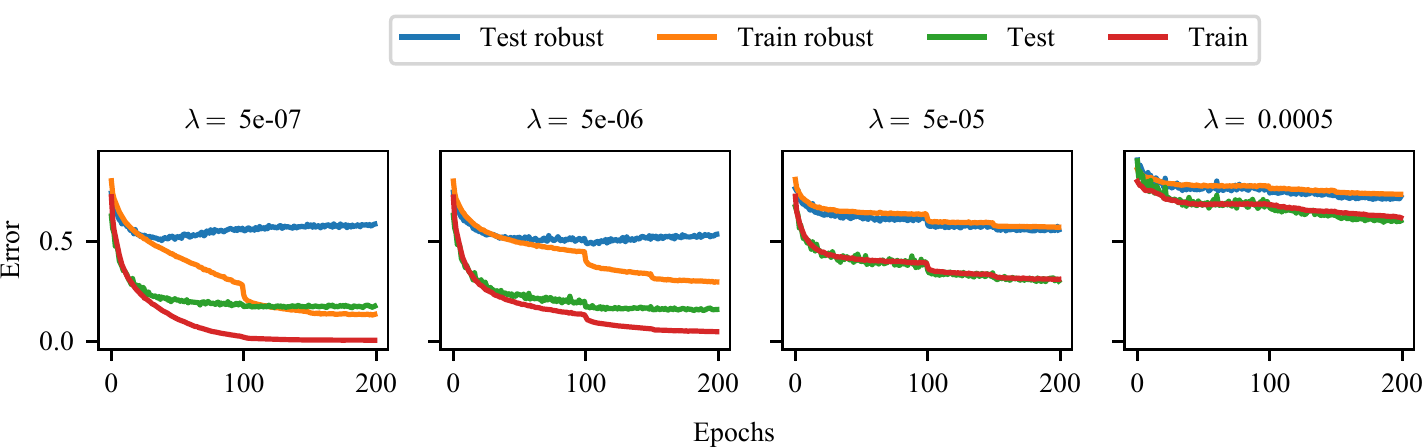}
\caption{Learning curves for adversarial training using $\ell_1$ regularization.}
\label{fig:l1_curves}
\end{figure*}

Figure \ref{fig:l1_curves} shows the corresponding learning curves for these four models. We see clear robust overfitting for the smaller two options in $\lambda$, and find no overfitting but highly regularized models for the larger two options, to the extent that there is no generalization gap and the training and testing curves actually appear to match.

\begin{figure*}[ht]
\centering
\includegraphics[scale=1]{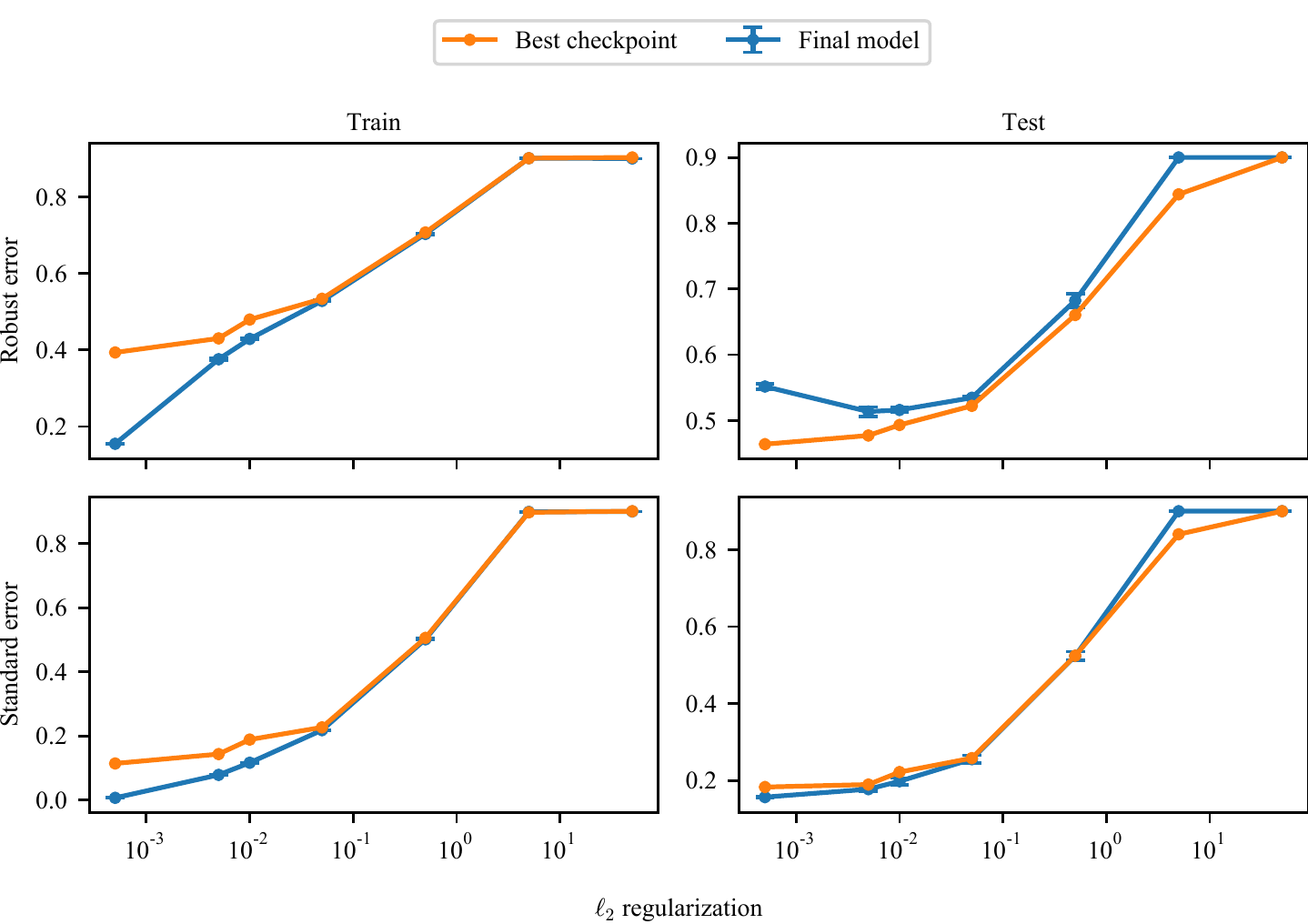}
\caption{Standard and robust performance on the train and test set using varying degrees of $\ell_2$ regularization.}
\label{fig:l2_generalization_all}
\end{figure*}

\begin{figure*}[ht]
\centering
\includegraphics[scale=1]{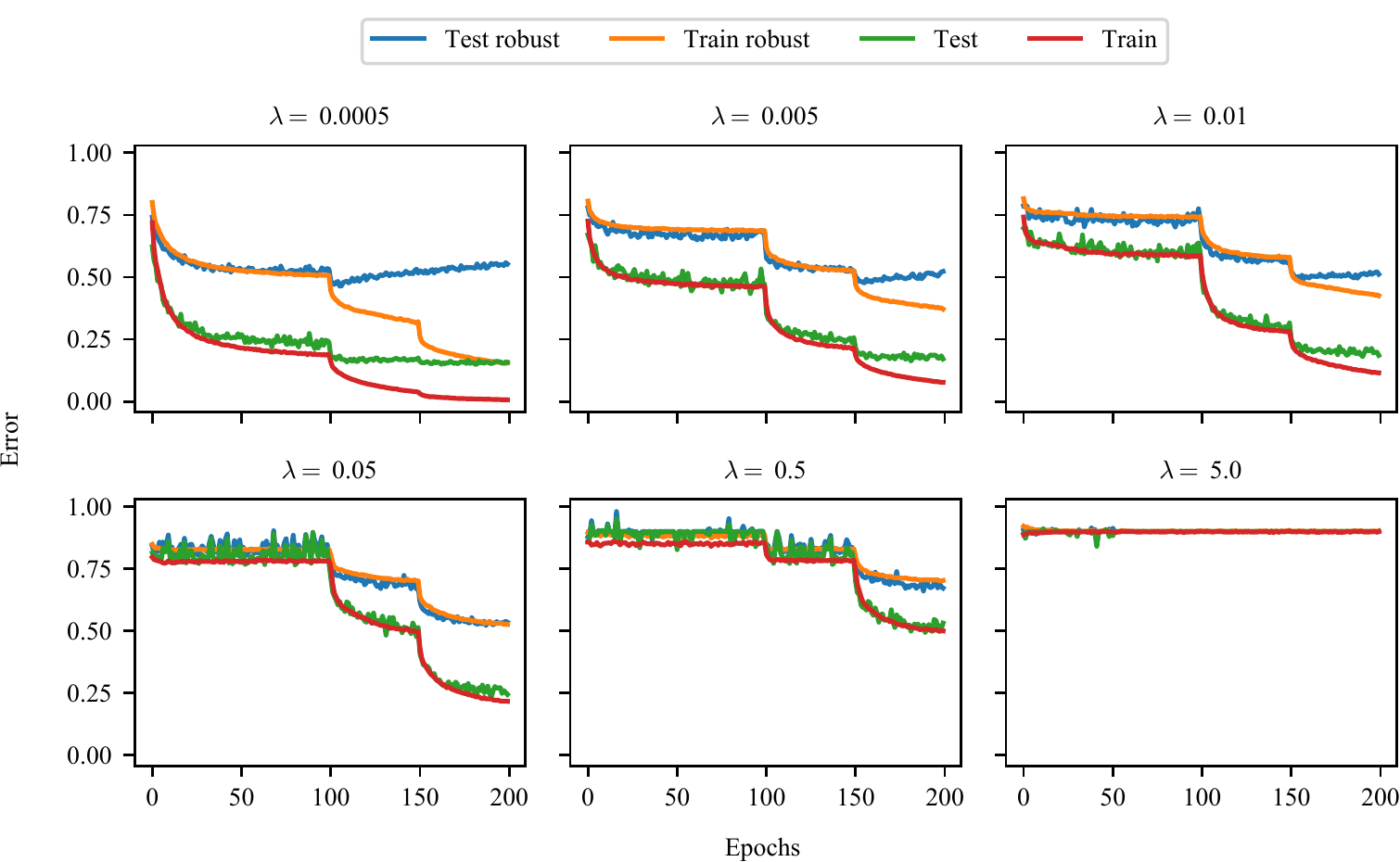}
\caption{Learning curves for adversarial training using $\ell_2$ regularization.}
\label{fig:l2_curves}
\end{figure*}

\paragraph{$\ell_2$ regularization}
Figure \ref{fig:l2_generalization_all} shows the training and testing performance of models using various degrees of $\ell_2$ regularization. We performed a search over regularization parameters $\lambda = \{5 \cdot 10^{k}\}$ for $k \in \{-4,-3,-2,-1,0\}$ as well as $\lambda = 0.01$. Note that $5 \cdot 10^{-4}$ is a fairly widely used value for weight decay in deep learning. We find that only the smallest choices
for $\lambda$ result in robust overfitting (e.g. $\lambda \leq 0.1$. However, inspecting the corresponding learning curves in Figure \ref{fig:l2_curves} reveals that the larger choices for $\lambda$ have a similar behavior to the larger forms of $\ell_1$ regularization, and end up with highly regularized models whose test performance perfectly matches the training performance at the cost of converging to a worse final robust test error.

\subsection{Data augmentation}
\label{app:data_aug}
In this section, we present additional details for the data augmentation approaches for preventing overfitting, namely cutout, mixup, and semi-supervised data.

\begin{figure*}[ht]
\centering
\includegraphics{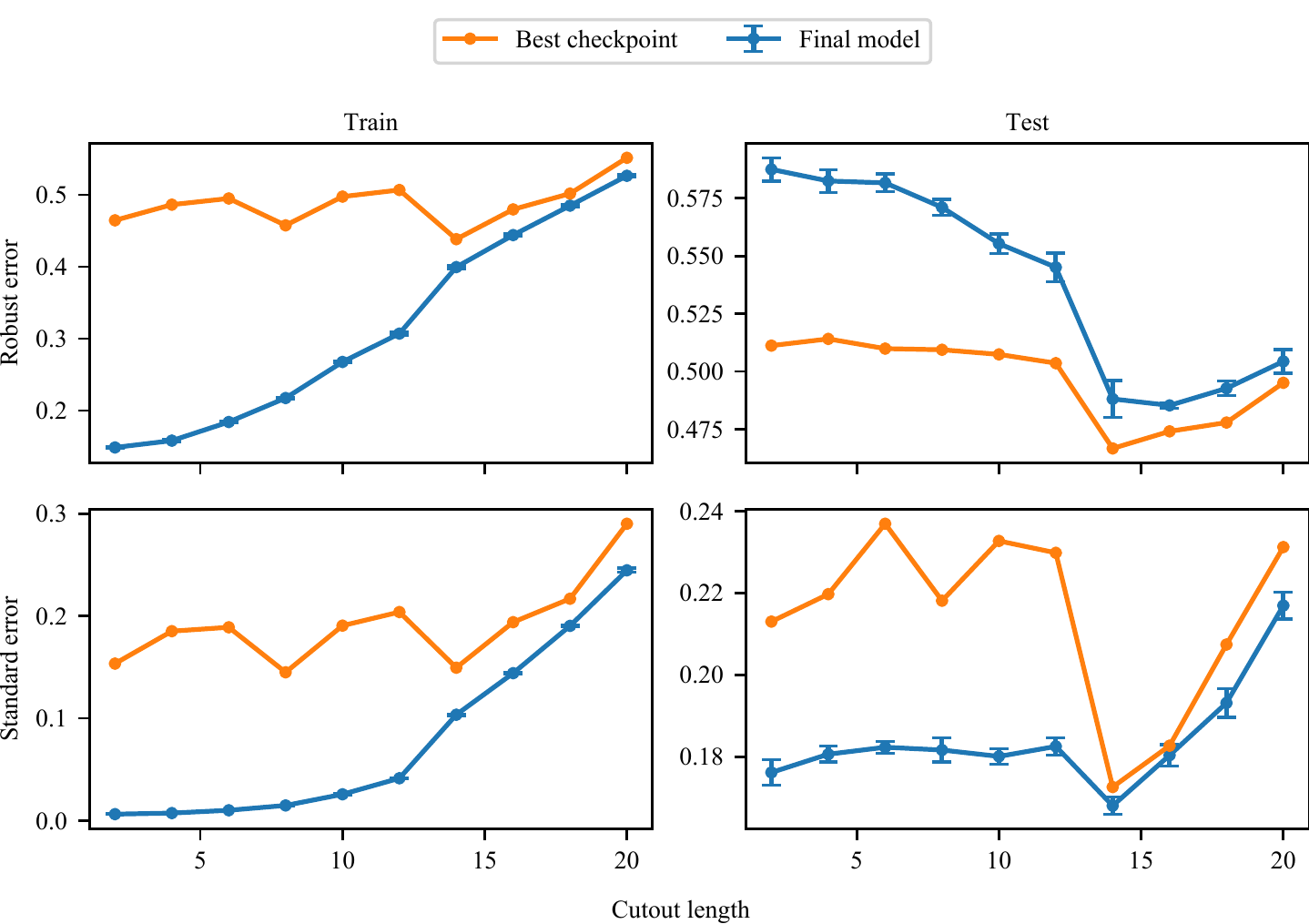}
\caption{Standard and robust performance on the train and test set for varying cutout patch lengths.}
\label{fig:cutout_generalization_all}
\end{figure*}

\begin{figure*}[ht]
\centering
\includegraphics{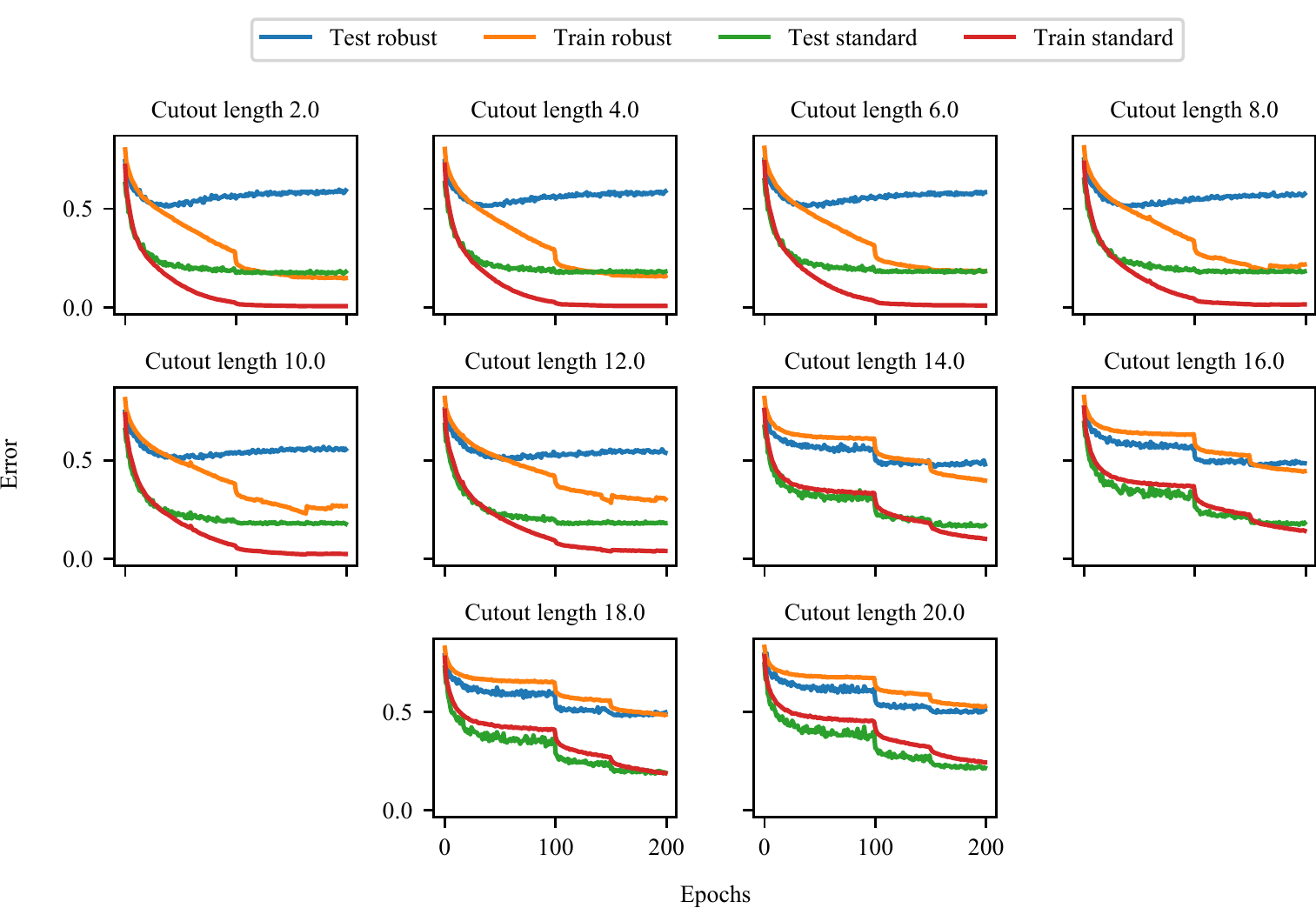}
\caption{Learning curves for adversarial training using cutout data augmentation with different cutout patch lengths.}
\label{fig:cutout_curves}
\end{figure*}

\paragraph{Cutout}
To analyze the effect of cutout on generalization, we range the cutout hyperparameter of patch length from 2 to 20. Figure \ref{fig:cutout_generalization_all} shows the training and testing performance of models using varying choices of patch lengths. Additionally, for each hyperparameter choice, we plot the resulting learning curves in Figure \ref{fig:cutout_curves}.

We find the optimal length of cutout patches to be 14, which on it's own is not quite as good as vanilla early stopping, but when combined with early stopping merely matches the performance of vanilla early stopping. In all cases, we observe robust overfitting to steadily degrade the robust test performance throughout training, with less of an effect as we increase the cutout patch length.

\begin{figure*}[ht]
\centering
\includegraphics{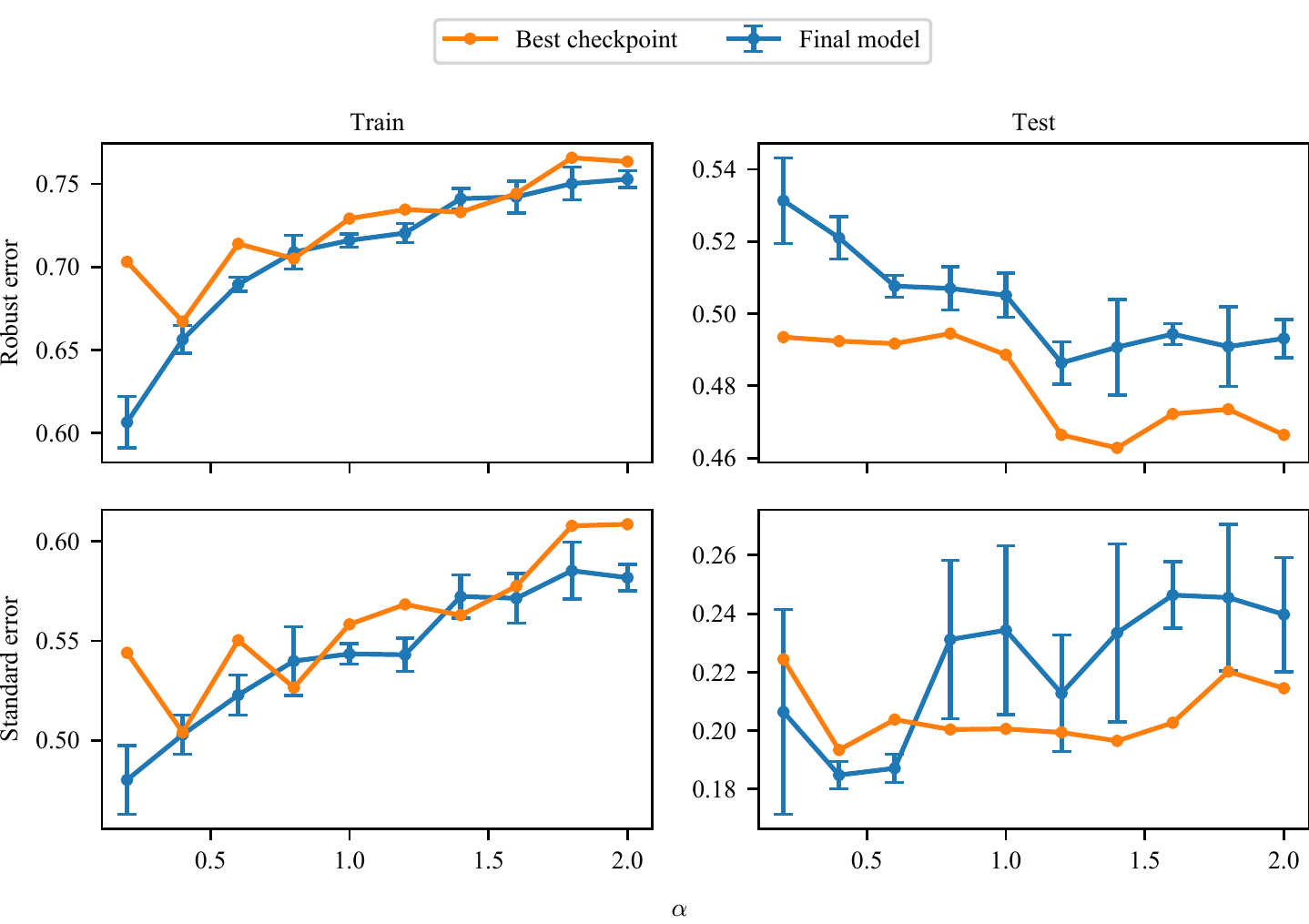}
\caption{Standard and robust performance on the train and test set for varying degrees of mixup.}
\label{fig:mixup_generalization_all}
\end{figure*}

\begin{figure*}[ht]
\centering
\includegraphics{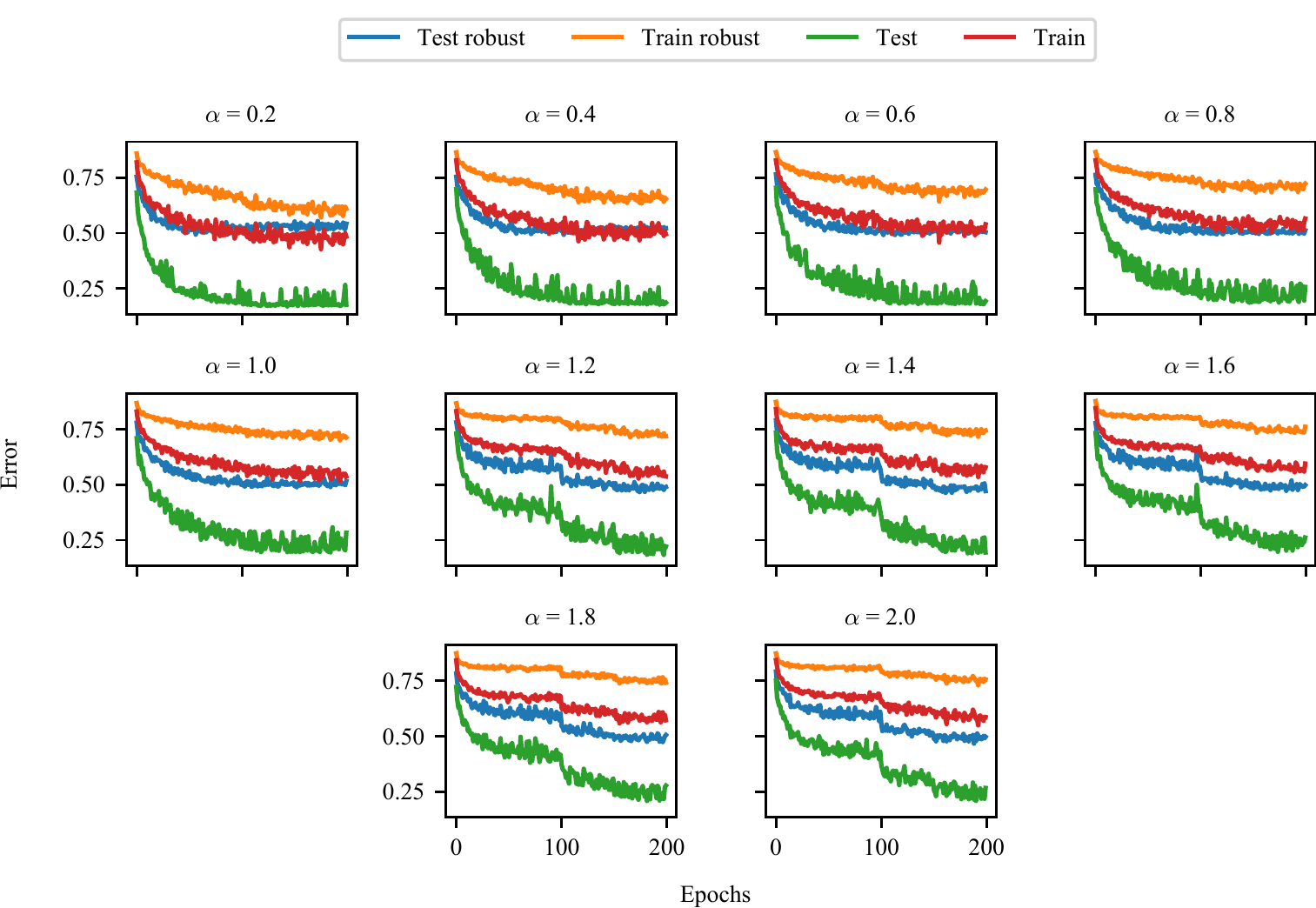}
\caption{Learning curves for adversarial training using mixup with different choices of hyperparameter $\alpha$.}
\label{fig:mixup_curves}
\end{figure*}

\paragraph{Mixup}
When training using mixup, we vary the hyperparameter $\alpha$ from 0.2 to 2.0. The training and testing performance of models using varying degrees of mixup can be found in Figure \ref{fig:mixup_generalization_all}. The resulting learning curves for each choice of $\alpha$ can be found in Figure \ref{fig:mixup_curves}.

For mixup, we find an optimal parameter value of $\alpha=1.4$. Similar to cutout, when combined with early stopping, it can only attain similar performance to vanilla early stopping, and otherwise converges to a worse model. However, although the learning curves for mixup training are significantly noisier than other methods, we do observe the robust test error to steadily decrease over training, indicating that mixup does stop robust overfitting to some degree (but does not obtain significantly better performance).

\section{Semi-supervised approaches}
\label{app:semisupervised}
For semi-supervised training, we use a batch size of 128 with equal parts labeled CIFAR-10 data and pseudo-labeled TinyImages data, as recommended by \citet{carmon2019unlabeled}. Each epoch of training is now equivalent in computation to two epochs of standard adversarial training. Note that the pre-activation ResNet18 is a smaller architecture than used by \citet{carmon2019unlabeled}, and so in our reproduction, the best checkpoint which achieves \cifarsemibest{}\% error is about 2\% higher than 38.5\%, which is what \citet{carmon2019unlabeled} can achieve with a Wide ResNet. Note that in the typical adversarially robust setting without additional semi-supervised data, a Wide ResNet can achieve about 3.5\% lower error than a pre-activation ResNet18.

We observe that the semi-supervised approach does not exhibit severe robust overfitting, as the smoothed learning curves tend to be somewhat relatively flat and don't show significant increases in robust test error. However, relative to the base setting of using only the original dataset, the robust test performance is extremely variable, with a range spanning almost 10\% robust error even when training error is relatively flat and has converged. As a result, it is critical to still use the best checkpoint even without robust overfitting, in order to avoid the fluctuations in test performance induced by the augmented training data.


%
%

\end{document}